\documentclass[]{IAC_style_2026}
\usepackage{xcolor}

\usepackage{placeins}
\begin{document}

\IACconference{77}
\IACpaperyear{2026}
\IACpapernumber{IAC-26-B2-IP-3-106854}
\IAClocation{Antalya, T\"urkiye}
\IACdate{5--9 October 2026}

\IACcopyrightB{Toshiki Tanaka}

\title{Enhancing Lunar PNT Performance Through Surface Station--Enabled Radiometric Observables}

\IACauthor{Toshiki Tanaka\textsuperscript{\orcidlink{0000-0002-1681-9155}}}{1}{0}
\IACauthor{Satoshi Ikari\textsuperscript{\orcidlink{0000-0002-0467-768X}}}{1}{1}

\IACauthoraffiliation{Intelligent Space Systems Laboratory, The University of Tokyo, Tokyo, Japan. \normalfont{E-mail:~\authormail{toshiki.tanaka@space.t.u-tokyo.ac.jp}}}

\abstract{Future lunar surface missions will require reliable and accurate positioning, particularly for autonomous rover operations. Although dedicated lunar navigation satellite systems are being developed, their vertical positioning performance can be limited by satellite availability and observation geometry, especially during early deployment phases. In this context, this study investigated a radiometric navigation architecture in which a surface station (SS) complements lunar navigation satellites by providing satellite observations for differential processing and an additional surface-to-surface ranging measurement to improve the vertical positioning performance of lunar surface users. The SS receives pseudorange and Doppler signals from the navigation satellites, which are used by the user to form double-differenced measurements that reduce clock and common-mode ephemeris errors and enable positioning of the rover relative to the SS. In addition, dual one-way ranging (DOWR) is performed between the SS and rover. The user (rover) then integrates the position solution obtained from these radiometric measurements with dead-reckoning information from a rover navigation sensor. Numerical Monte Carlo simulations were performed using five lunar navigation satellites representing NASA's LCRNS constellation and a rover trajectory around Connecting Ridge near the lunar south pole. The results showed that introducing the SS improved both horizontal and vertical relative positioning performance, and that SS--rover ranging provided an additional improvement. In the simulated representative case, the mean 99.7\% vertical positioning error was reduced from 29.9~m without the SS to 5.5~m when the SS was used for double-differenced pseudorange and Doppler processing, and further reduced to 4.8~m when SS--rover DOWR was additionally included. The simulations also showed that an uncorrected DOWR hardware bias could significantly degrade positioning performance, whereas the proposed bias-estimation process reduced the final DOWR-bias estimation error to $0.058 \pm 0.057$~m. The results further revealed that the benefit of SS--rover ranging depended on the observation geometry. Overall, the results quantitatively demonstrated the potential of a lunar SS to play an additional role beyond communication, differential corrections, and satellite orbit determination: directly improving the vertical relative positioning of lunar surface users through an additional radiometric ranging measurement.}

\maketitle
\thispagestyle{fancy}


\section{Introduction}

As lunar exploration enters a new phase of sustained and increasingly
complex surface activities, the demand for reliable navigation
capabilities on the lunar surface is rapidly growing. In particular,
autonomous navigation has been identified as an important capability
for future lunar surface missions \cite{marine-roig2026lunar}.

Several international efforts are currently developing dedicated
lunar communication and navigation infrastructures, including NASA's
Lunar Communications Relay and Navigation Systems (LCRNS), ESA's
Moonlight programme, and JAXA's Lunar Navigation Satellite System
(LNSS), with the broader goal of providing interoperable positioning,
navigation, timing, and communication (PNT+C) services around the Moon
\cite{nasa2024lcrns,esa2022moonlight,murata2022lnss,
nasa2023lunanet,jun2025rapid}. Such architectures are expected to
eventually provide Global Navigation Satellite System (GNSS)-like
services to a broad range of lunar surface users. However, deployment
of a sufficiently large lunar constellation requires substantial
infrastructure investment and can take many years, while the number of
available orbiters and the resulting navigation geometry may remain
limited during the early deployment phase \cite{jun2025rapid}.

As a means of improving navigation system performance, a surface
station (SS) with a known location can complement the orbital
infrastructure by providing additional navigation functions to users
on the lunar surface and in the vicinity of the Moon. In previous
studies, SS have primarily been considered for two purposes. First,
measurements obtained at known SS locations can be used to generate
differential corrections and mitigate common ranging error sources
\cite{marine-roig2026lunar,cheung2024ground,
akiyama2022differential,jun2025rapid}. Second, SS-originated ranging
can support orbit determination of lunar navigation satellites
\cite{cheung2024ground,iiyama2019optimization}.

The potential role of SS, however, is not limited to these
applications. This study investigates a third use of SS: providing
additional ranging signals to improve user positioning accuracy,
particularly in the vertical direction (the local Up direction in the
topocentric coordinate frame).

Current lunar navigation satellite constellation designs primarily
focus on horizontal user positioning performance, while explicit
requirements for vertical positioning accuracy have not been clearly
established \cite{nasa2022lunarrelay,murata2022lnss}. Therefore,
ranging signals from these lunar navigation satellites could result
in relatively weak observability in the vertical direction. To this
end, a SS-originated ranging measurement provides a different
line-of-sight geometry, improving measurement diversity and thereby
the vertical positioning accuracy of the rover.

The use of surface-to-surface ranging for positioning is not new.
Terrestrial applications of surface-to-surface ranging have been
extensively investigated using technologies including ultra-wideband
(UWB), for purposes such as cooperative positioning
\cite{deponte2017survey}. Similar concepts have also been investigated
for lunar surface navigation
\cite{pohlmann2025hybrid,kim2026conceptual}, including the CADRE flight
mission \cite{knowles2024multirobot}. However, previous lunar studies
have mainly focused on horizontal positioning or general relative
localization. To the best of the authors' knowledge, this is the first
study to specifically evaluate the \textit{vertical} positioning benefit of
SS-originated ranging for lunar surface users.

Alternatively, the vertical position can be constrained using a lunar
digital elevation model (DEM). A lunar DEM represents the lunar surface
topography and provides terrain elevation information. Existing lunar
DEMs are constructed from satellite remote-sensing observations rather
than in-situ measurements on the lunar surface. Using such DEM
information, DEM-based altitude constraints have previously been
incorporated into lunar rover navigation and positioning algorithms
\cite{sabatini2025satellite,audet2024positioning,melman2022lcns}.

In these methods, the achievable vertical positioning accuracy depends directly on the accuracy of the lunar DEM. For example, SLDEM2015, which covers latitudes within $\pm60^\circ$, has a typical vertical accuracy of approximately 3--4 m \cite{barker2016dem}. Higher-resolution and more accurate DEMs have been developed for selected regions, particularly near the lunar poles, such as NASA PGDA's High-Resolution LOLA Topography for Lunar South Pole Sites product \cite{Barker2021}, while Bertone et al. \cite{Bertone2026} showed that the effective resolution of LOLA-based polar DEMs varies with observation density, leaving some terrain features unresolved between LOLA tracks. However, such high-resolution and high-accuracy DEMs are available only for selected regions of the lunar surface. Furthermore, existing lunar DEM accuracy assessments have primarily relied on internal consistency or cross-comparison among remotely sensed datasets, rather than direct comparison with ground-truth elevation measurements on the lunar surface. This limitation further motivates the development of a vertical positioning method that does not rely on a lunar DEM.

Therefore, this study evaluates a radiometric navigation architecture
that does not rely on a DEM-derived altitude constraint, with the goal
of achieving high vertical positioning accuracy. The proposed
architecture consists of lunar navigation satellites, a SS, and a
surface user. The SS receives radiometric signals from the
navigation satellites and uses these measurements to generate
differential corrections for the user. In addition, the SS provides a
direct radiometric measurement to the user through dual one-way ranging
(DOWR). The user (rover) then combines the satellite-based measurements, the
differential information provided by the SS, the SS--rover DOWR
measurement, and its navigation-sensor measurements through filter-based
estimation.

Based on this architecture, this study investigates the error sources
inherent in the proposed navigation scheme, develops methods to mitigate
or estimate these errors, and evaluates the resulting benefits and
limitations of the architecture through numerical simulations.
The simulations used five lunar navigation satellites based on the
latest available LCRNS orbital configuration
\cite{ryden2025lcrns}. The SS location and rover trajectory were defined
around Connecting Ridge, one of the candidate regions for Artemis lunar
south-pole exploration \cite{gruener2020shackleton}.

In summary, the main contributions of this paper are as follows:
\begin{itemize}
\item \textbf{Architecture design and error mitigation strategy development:}
A SS-enabled navigation architecture is proposed, and key error sources
and their mitigation methods are identified and investigated.

\item \textbf{Performance benchmarking:}
Numerical simulations evaluate the vertical positioning performance,
limitations, and applicability of the proposed architecture.
\end{itemize}


\section{System Architecture}

\begin{figure*}[!ht]
\centering
\includegraphics[
    width=0.85\linewidth,
    trim={{.0\linewidth} 0 0 0},
    clip
]{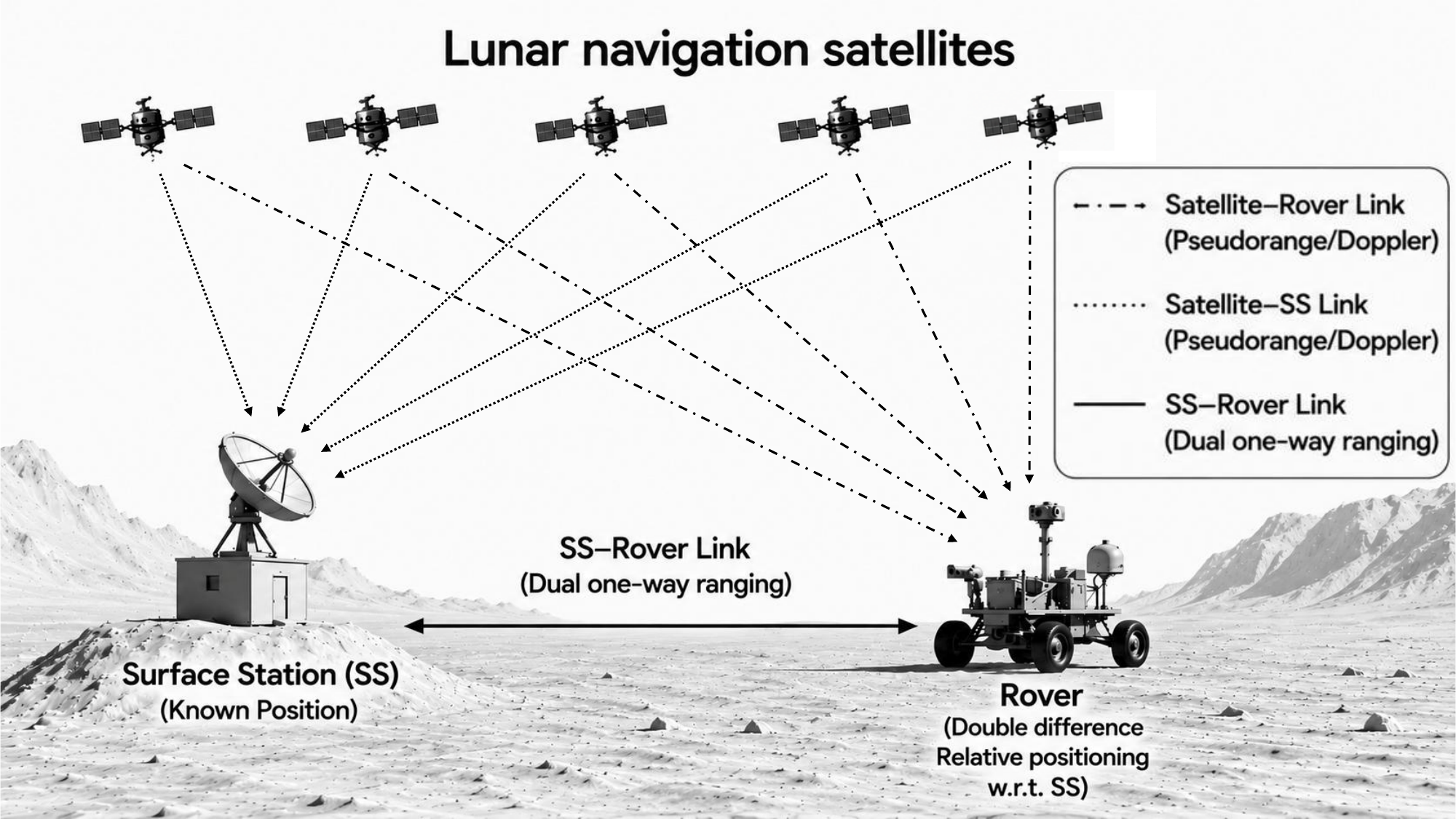}
\caption{\footnotesize Overview of the radiometric architecture. (The
figure was generated using the image generation feature in ChatGPT
GPT-5.6 Sol and subsequently edited by the author.)}
\label{CONOPS}
\end{figure*}

The proposed navigation architecture uses radiometric ranging signals
from a LCRNS satellite constellation together with signals from a SS
installed at a known location on the lunar surface. The SS--rover
ranging signal is assumed to be available only when LOS between the SS
and rover is maintained, considering possible terrain blockage. The
radiometric architecture is illustrated in Fig.~\ref{CONOPS}.

The rover is also assumed to use an inertial navigation system (INS),
and therefore the required navigation accuracy does not need to be
achieved by radiometric ranging alone. Instead, radiometric measurements
and INS measurements are integrated through sensor fusion. The two
measurement sources are complementary because they have different error
characteristics. In particular, when satellite geometry is unfavorable
and radiometric positioning performance degrades, the INS can be used
to propagate the rover state until reliable radiometric measurements
become available again. Similar integration of radiometric measurements
and onboard navigation sensors has been considered in previous lunar
rover navigation studies
\cite{sabatini2025satellite,melman2022lcns}.

The types of radiometric measurements used in the proposed architecture
are selected considering several representative challenges associated
with lunar navigation. The following issues describe these challenges
and how they are addressed in the proposed architecture.

\textbf{Issue 1: Integration of Pseudorange and Doppler Measurements}

Current lunar navigation satellite systems are designed to provide
pseudorange measurements as a primary ranging observable for user
positioning. In addition to pseudorange measurements, Doppler
measurements from the lunar navigation satellites are assumed to be
available at both the SS and rover in this study. Doppler provides
independent range-rate information and therefore increases the available
measurement information.

\textbf{Issue 2: Satellite clock, satellite ephemeris, and lunar ephemeris errors}

Satellite clock and ephemeris errors are important error sources in
satellite-based navigation. In terrestrial GNSS, these parameters can
be estimated and updated with high accuracy using frequent observations
from a large network of ground stations. For lunar navigation systems,
however, the long Earth--Moon distance and the limited number of ground
stations available for lunar missions may result in larger uncertainties
in these parameters. The proposed architecture therefore assumes that
satellite clock and ephemeris errors cannot always be neglected.

To reduce their impact, the proposed architecture uses double difference
(DD) measurements between the SS and rover. DD is an established
differential positioning technique that combines measurements from two
receivers and two satellites to eliminate satellite and receiver clock
errors and mitigate common-mode components of satellite ephemeris
errors, enabling accurate positioning of the rover relative to the SS
\cite{tanaka2026single}. The effectiveness of differential positioning
in reducing common-mode errors depends on the SS--rover baseline and
measurement geometry. In general, stronger common-mode error cancellation
can be expected when the two receivers are sufficiently close relative
to the satellite range. In this study, the SS--rover baseline ranged
up to approximately 6 km.

In addition, other common errors associated with the lunar reference
frame and ephemerides can also be reduced by relative positioning.

\textbf{Issue 3: SS clock error and time transfer}

The clock error of the SS is another potential error source. If a
conventional one-way pseudorange measurement is used between the SS and
rover, the relative clock offset directly appears in the ranging
measurement. To reduce this dependency, DOWR is used for the SS--rover
link. By combining one-way measurements in both directions, DOWR
removes the dominant relative clock offset between the SS and rover
\cite{tanaka2026single}. A similar approach is two-way ranging, which
also does not require precise clock synchronization between the SS and
rover and could alternatively be adopted in place of DOWR.

\textbf{Issue 4: DOWR hardware bias}

Although DOWR reduces the effect of the relative clock offset between
the SS and rover, the measurement can contain hardware-dependent biases
caused by signal delays in the transmitting and receiving hardware
\cite{tanaka2026single}. Internal loopback measurements can determine
the combined transmit (TX) and receive (RX) delays at each node, but
cannot separately determine the individual TX and RX delays. Therefore,
a residual bias can remain in the DOWR measurement.

In this study, the residual DOWR hardware bias is estimated and removed
through filtering. The observability and convergence of the DOWR bias
estimation depend on the available measurement geometry and rover motion
and were therefore evaluated through numerical simulations.

\textbf{Issue 5: SS position error}

An error in the assumed SS position affects absolute rover positioning.
In the proposed architecture, however, the rover position is estimated
relative to the SS, and therefore the absolute SS position error does
not directly affect the SS-relative rover position.


\section{Mathematical Models}

\subsection{Radiometric Observation}

The proposed navigation architecture uses three types of radiometric
observations: pseudorange and Doppler measurements from the lunar
navigation satellites, and DOWR measurements between the SS and rover.

\subsubsection{Pseudorange and Doppler Observations}

The pseudorange measurement between receiver $u$ and navigation
satellite $s_i$ at time $t$ is modeled as
\cite{tanaka2026single}
\begin{equation}
\rho_u^{s_i}(t)
=
r_u^{s_i}(t)
+
c\left(
\delta t_u(t)-\delta t^{s_i}(t)
\right)
+
\epsilon_{\rho,u}^{s_i}(t)
\end{equation}
where the geometric range $r_u^{s_i}$ is given by
\begin{equation}
r_u^{s_i}(t)
=
\left\|
\mathbf{x}^{s_i}(t)-\mathbf{x}_u(t)
\right\|.
\end{equation}
Here, $s_i$ denotes the $i$th navigation satellite,
$\rho_u^{s_i}$ is the measured pseudorange between receiver $u$
and satellite $s_i$, $\mathbf{x}^{s_i}$ and $\mathbf{x}_u$ are
the satellite and receiver positions, respectively, $c$ is the speed
of light, $\delta t^{s_i}$ and $\delta t_u$ are the satellite
and receiver clock biases, respectively, and
$\epsilon_{\rho,u}^{s_i}$ represents the pseudorange measurement
noise. The pseudorange measurement noise is modeled as zero-mean
Gaussian noise with standard deviation $\sigma_{r,\mathrm{PR}}$.

In the numerical simulations, the geometric range was calculated by
accounting for the motion of both the satellite and receiver, as well
as lunar rotation, during the signal propagation time. Thus, the
Sagnac effect associated with the Moon-fixed reference frame was
included in the simulated pseudorange observations.

The Doppler measurement is expressed in terms of the range rate as
\cite{tanaka2026single}
\begin{equation}
f_{\mathrm{Doppler},u}^{s_i}(t)
=
-\frac{f_c}{c}
\dot{r}_u^{s_i}(t)
\end{equation}
where $f_c$ is the carrier frequency and $\dot{r}_u^{s_i}$ is the
geometric range rate between satellite $s_i$ and receiver $u$.
Considering clock-related and measurement errors, the corresponding
Doppler observation is expressed as
\begin{equation}
\tilde{f}_{\mathrm{Doppler},u}^{s_i}(t)
=
-\frac{f_c}{c}
\dot{\rho}_u^{s_i}(t),
\label{eq:doppler_observation}
\end{equation}
where $\dot{\rho}_u^{s_i}$ denotes the pseudorange rate.

In this study, both the SS and rover receive pseudorange and Doppler
measurements from the lunar navigation satellites. To reduce common
error sources, the rover navigation filter uses DD
observations between the rover, SS, and navigation satellites.
Here, $\Delta\nabla$ denotes the DD operator between
two receivers and two satellites. Using satellite $s_j$ as the
reference satellite, the DD pseudorange for the $i$th
satellite $s_i$ is defined as
\begin{equation}
\Delta\nabla\rho^{s_i,s_j}(t)
=
\left(
\rho_R^{s_i}(t)-\rho_{SS}^{s_i}(t)
\right)
-
\left(
\rho_R^{s_j}(t)-\rho_{SS}^{s_j}(t)
\right)
\end{equation}
where the subscripts $R$ and $SS$ denote the rover and SS,
respectively. The receiver and satellite clock biases are removed by
the DD operation.

The effect of satellite ephemeris errors can also be reduced through
differential positioning. Let the ephemeris-error-perturbed position
of satellite $s_i$ be expressed as \cite{tanaka2026single}
\begin{equation}
\hat{\mathbf{x}}^{s_i}(t)
=
\mathbf{x}^{s_i}(t)
+
\delta\mathbf{x}_{\mathrm{eph}}^{s_i}(t),
\label{eq:satellite_ephemeris_error}
\end{equation}
where $\delta\mathbf{x}_{\mathrm{eph}}^{s_i}$ denotes the satellite
ephemeris error. By linearizing the geometric range with respect to the
satellite position, the ephemeris-induced error remaining after single
differencing between the rover and SS is approximately
\begin{equation}
\delta\mathbf{x}_{\mathrm{eph}}^{s_i}(t)
\cdot
\left(
\mathbf{e}_{R}^{s_i}(t)
-
\mathbf{e}_{SS}^{s_i}(t)
\right).
\end{equation}
When the SS--rover baseline is sufficiently short compared with the
satellite range, the corresponding line-of-sight unit vectors are
similar, and therefore the effect of the satellite ephemeris error is
largely canceled \cite{tanaka2026single}. Consequently, double
differencing removes the receiver and satellite clock biases while
substantially reducing the effect of satellite ephemeris errors.

Similarly, the DD pseudorange-rate observation is
defined as
\begin{equation}
\Delta\nabla\dot{\rho}^{s_i,s_j}(t)
=
\left(
\dot{\rho}_R^{s_i}(t)-\dot{\rho}_{SS}^{s_i}(t)
\right)
-
\left(
\dot{\rho}_R^{s_j}(t)-\dot{\rho}_{SS}^{s_j}(t)
\right).
\end{equation}
The corresponding DD Doppler observation can therefore
be expressed as
\begin{equation}
\Delta\nabla\tilde{f}_{\mathrm{Doppler}}^{s_i,s_j}(t)
=
-\frac{f_c}{c}
\Delta\nabla\dot{\rho}^{s_i,s_j}(t).
\end{equation}

\subsubsection{DOWR}

For the surface-to-surface link between the SS and rover, DOWR is
used to obtain a ranging measurement without requiring precise clock
synchronization between the two nodes. In DOWR, each node transmits a
ranging signal to the other node, and the transmission and reception
times are measured using the local clock at each node. The mathematical
model of DOWR used in this study is based on
Ref.~\cite{tanaka2026single}.

Let $t_{SS,1}$ and $t_{R,1}$ denote the transmission time at the SS
and reception time at the rover for the \textit{SS-to-rover} signal,
respectively. Similarly, let $t_{R,2}$ and $t_{SS,2}$ denote the
transmission time at the rover and reception time at the SS for the
\textit{rover-to-SS} signal. The DOWR observation is then defined as
\begin{equation}
\rho_{\mathrm{DOWR}}
=
\frac{c}{2}
\left[
(t_{R,1}-t_{SS,1})
+
(t_{SS,2}-t_{R,2})
\right].
\label{eq:dowr_definition}
\end{equation}

Because DOWR is formed from bidirectional transmission and reception
events occurring at different times, a single continuous time $(t)$ is
not assigned to the DOWR observation. Assuming that the clock drift during the signal propagation is
negligible, by combining the two one-way measurements, the constant
relative clock bias between the SS and rover is eliminated through the
subtractions $t_{R,1}-t_{R,2}$ and $t_{SS,2}-t_{SS,1}$ in
Eq.~\ref{eq:dowr_definition}.

Assuming that the change in the SS--rover geometric range during the
DOWR exchange is negligible, the DOWR observation can be approximated as
\begin{equation}
\tilde{\rho}_{\mathrm{DOWR}}
=
\left\|
\mathbf{x}_{R}
-
\mathbf{x}_{SS}
\right\|
+
b_{\mathrm{DOWR}}
+
\epsilon_{\mathrm{DOWR}},
\label{eq:dowr_observation}
\end{equation}
where $b_{\mathrm{DOWR}}$ denotes the residual hardware-dependent bias
in the DOWR measurement and $\epsilon_{\mathrm{DOWR}}$ denotes the
DOWR measurement noise. The DOWR measurement noise is modeled as
zero-mean Gaussian noise with standard deviation
$\sigma_{r,\mathrm{DOWR}}$.

The DOWR hardware bias $b_{\mathrm{DOWR}}$ affects the user positioning
accuracy. To mitigate this effect, the proposed architecture estimates the DOWR
hardware bias during the radiometric position estimation and subsequently removes
it from the DOWR measurements. Its effect on the positioning performance was
evaluated in the numerical simulations.

\subsubsection{Integrated Radiometric Positioning and DOWR Bias Estimation}

The DOWR hardware bias is estimated jointly with the rover position
during the radiometric estimation process. The proposed estimator combines
iterative Gauss--Newton relinearization with a final Kalman filter (KF)
update, allowing nonlinear measurement geometry to be handled while
retaining covariance-based state estimation. This implementation follows
the Gauss--Newton interpretation of the iterated Kalman filter update
described by Bell and Cathey \cite{bell1993iterated}.

When DOWR-bias estimation is performed, the radiometric correction state
is augmented as
\begin{equation}
\delta\mathbf{x}_{\mathrm{rad},k}^{\mathrm{aug}}
=
\begin{bmatrix}
\delta\mathbf{x}_{R,k}^{T} &
\delta b_{\mathrm{DOWR},k}
\end{bmatrix}^{T},
\end{equation}
where $\delta\mathbf{x}_{R,k}$ is the rover-position correction and
$\delta b_{\mathrm{DOWR},k}$ is the correction to the current DOWR-bias
estimate. Hereafter, $k$ denotes the estimation epoch.

At iteration $\ell$, the radiometric residual vector and Jacobian are
evaluated at the current linearization point as
\begin{equation}
\boldsymbol{\nu}_{k}^{(\ell)}
=
\mathbf{z}_{k}
-
\mathbf{h}
\left(
\hat{\mathbf{x}}_{\mathrm{rad},k}^{(\ell-1)}
\right),
\end{equation}
and
\begin{equation}
\mathbf{H}_{k}^{(\ell)}
=
\left.
\frac{\partial\mathbf{h}}
{\partial\mathbf{x}_{\mathrm{rad}}}
\right|_{\hat{\mathbf{x}}_{\mathrm{rad},k}^{(\ell-1)}},
\label{eq:radiometric_jacobian}
\end{equation}
where $\mathbf{z}_{k}$ contains the active DD pseudorange, DD Doppler,
and, when available, SS--rover DOWR observations, and
$\mathbf{h}(\cdot)$ is the nonlinear radiometric measurement function
that predicts these observations from the radiometric state. When the
DOWR-bias state is inactive, $\mathbf{H}_{k}^{(\ell)}$ contains only
the rover-position state columns.

The damped Gauss--Newton correction is calculated as
\begin{equation}
\begin{aligned}
\Delta\mathbf{x}_{\mathrm{rad},k}^{(\ell)}
&=
\left[
\left(\mathbf{H}_{k}^{(\ell)}\right)^{T}
\mathbf{H}_{k}^{(\ell)}
+
\lambda_{k}^{(\ell)}\mathbf{I}
\right]^{-1}
\left(\mathbf{H}_{k}^{(\ell)}\right)^{T}
\boldsymbol{\nu}_{k}^{(\ell)},
\\
&\qquad \ell=1,\ldots,4.
\end{aligned}
\label{eq:damped_gauss_newton}
\end{equation}
The damping parameter $\lambda_{k}^{(\ell)}$ is adaptively scaled, with
larger damping applied under poorer three-dimensional observation
geometry.

For the final KF update, the predicted covariance and Kalman gain are
given by
\begin{equation}
\mathbf{P}_{k}^{-}
=
\mathbf{P}_{k-1}^{+}
+
\mathbf{Q},
\end{equation}
and
\begin{equation}
\mathbf{K}_{k}
=
\mathbf{P}_{k}^{-}
\left(\mathbf{H}_{k}^{(5)}\right)^{T}
\left[
\mathbf{H}_{k}^{(5)}
\mathbf{P}_{k}^{-}
\left(\mathbf{H}_{k}^{(5)}\right)^{T}
+
\mathbf{R}_{k}
\right]^{-1}.
\end{equation}
The final correction is
\begin{equation}
\Delta\hat{\mathbf{x}}_{\mathrm{rad},k}^{\mathrm{aug}}
=
\mathbf{K}_{k}
\boldsymbol{\nu}_{k}^{(5)}.
\end{equation}
The rover-position estimate is updated as
\begin{equation}
\hat{\mathbf{x}}_{R,k}^{+}
=
\hat{\mathbf{x}}_{R,k}^{-}
+
\Delta\hat{\mathbf{x}}_{R,k}.
\end{equation}
The retained bias estimate is updated as
\begin{equation}
\hat{b}_{\mathrm{DOWR},k}^{+}
=
\hat{b}_{\mathrm{DOWR},k}^{-}
+
\Delta\hat{b}_{\mathrm{DOWR},k}.
\end{equation}
Finally, the covariance is updated using
\begin{equation}
\begin{aligned}
\mathbf{P}_{k}^{+}
={}&
\left(
\mathbf{I}
-
\mathbf{K}_{k}\mathbf{H}_{k}^{(5)}
\right)
\mathbf{P}_{k}^{-}
\left(
\mathbf{I}
-
\mathbf{K}_{k}\mathbf{H}_{k}^{(5)}
\right)^{T}
\\
&+
\mathbf{K}_{k}\mathbf{R}_{k}\mathbf{K}_{k}^{T}.
\end{aligned}
\end{equation}

During the preceding Gauss--Newton iterations, the DOWR-bias estimate
is temporarily updated as part of the augmented correction state for
relinearization. However, if the final KF update of the DOWR bias is
disabled based on the criterion described next, these temporary
Gauss--Newton bias updates are discarded, and the DOWR-bias estimate
remains unchanged from the last effective update.

\paragraph{Selective DOWR-Bias Update}
The PDOP-like metric $D_{P,k}$ is used to exclude DOWR-bias updates
under unfavorable observation geometry. The metric is calculated from
the final-iteration Jacobian as
\begin{equation}
D_{P,k}
=
\sqrt{
\sum_{j=1}^{3}
\left[
\left(
\left(\mathbf{H}_{k}^{(5)}\right)^{T}
\mathbf{H}_{k}^{(5)}
\right)^{-1}
\right]_{j,j}
}.
\label{eq:pdop_like_metric}
\end{equation}
where the first three diagonal elements correspond to the rover
position states in the East-North-Up (ENU) frame. Because the Jacobian
may include pseudorange, weighted Doppler, and SS--rover DOWR
observation rows with different physical units and weighting,
$D_{P,k}$ is referred to as a PDOP-like metric rather than a
conventional geometry-only PDOP.

The retained DOWR-bias estimate is updated in the final KF update only
when the SS--rover DOWR observation is available and $D_{P,k}$ satisfies
the prescribed DOWR-bias-update threshold.

\subsection{Sensor Fusion}

\begin{figure*}[!ht]
\centering
\includegraphics[
    width=0.98\linewidth,
    trim={0 0 0 0},
    clip
]{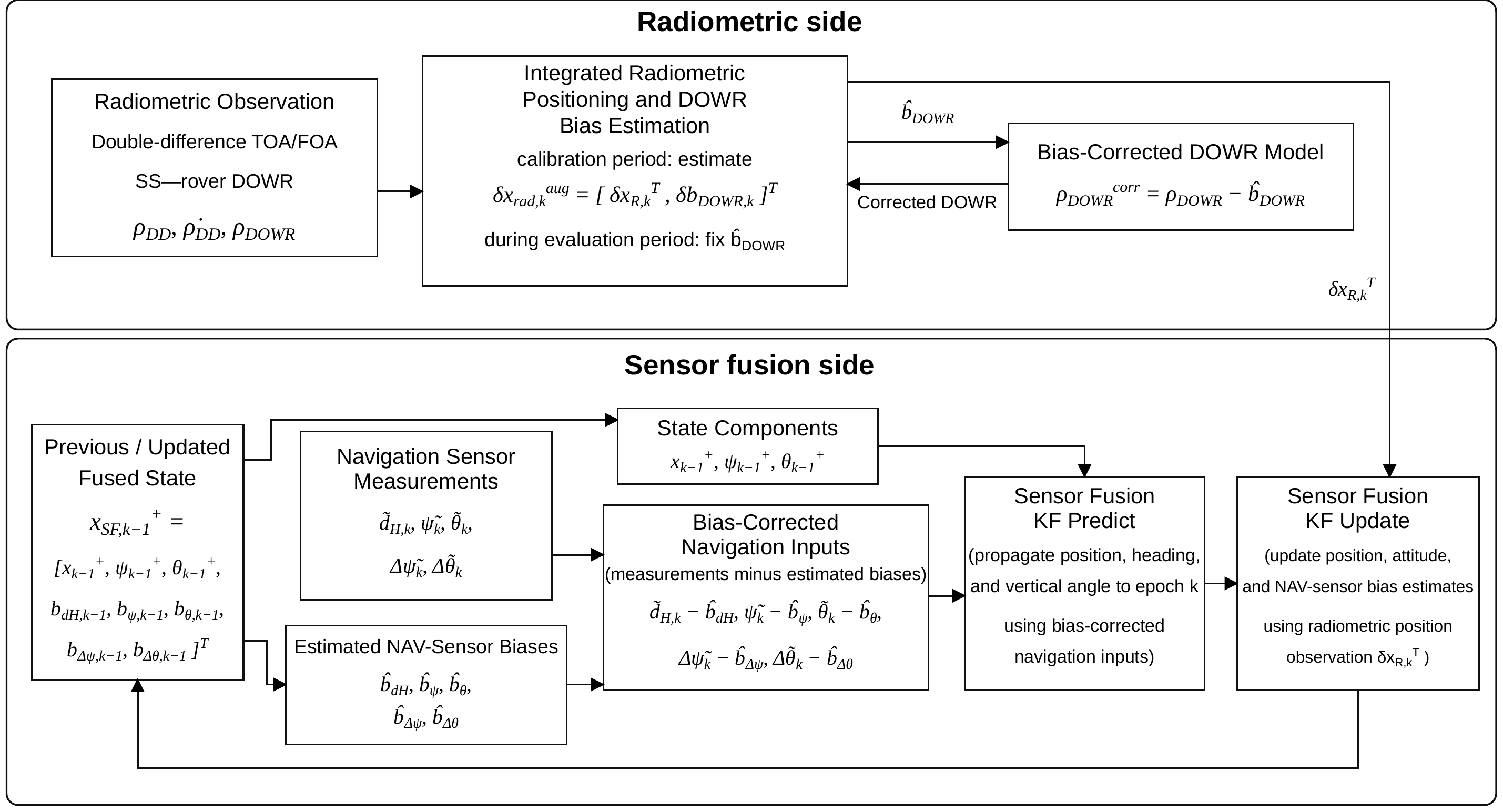}
\caption{\footnotesize Sensor Fusion Architecture. (The figure was
generated using the image generation feature in ChatGPT GPT-5.6 Sol
and subsequently edited by the author.)}
\label{MathModel}
\end{figure*}

The proposed navigation architecture combines radiometric positioning
with relative-motion measurements from a navigation sensor. The sensor
fusion flow is illustrated in Fig.~\ref{MathModel}.

The radiometric position solution is provided to the
sensor-fusion KF, which integrates it with the navigation-sensor
measurements to estimate the rover state and navigation-sensor biases.
When a valid radiometric position solution is unavailable, the
navigation-sensor measurements propagate the rover state until the next
radiometric update.

\subsubsection{Navigation Sensor Model}

In this study, the INS is represented by an abstracted navigation-sensor
model rather than raw inertial measurements. A detailed assessment of
the validity and limitations of this abstraction is beyond the scope of
the present study.

The navigation-sensor quantities used in this study are defined as
follows:
\begin{itemize}
    \item $\tilde{d}_{H,k}$ and $d_{H,k}$ denote the measured and true
    horizontal travel distances, respectively. The horizontal travel
    distance is defined as the projection of the three-dimensional rover
    displacement onto the local horizontal plane.

    \item $\tilde{\psi}_k$ and $\psi_k$ denote the measured and true
    horizontal heading angles, respectively.

    \item $\tilde{\theta}_k$ and $\theta_k$ denote the measured and true
    vertical angles, respectively.

    \item $\Delta\tilde{\psi}_k$ and $\Delta\tilde{\theta}_k$ denote the
    measured changes in horizontal heading and vertical angle between
    consecutive epochs, respectively.
\end{itemize}

Each navigation-sensor measurement is modeled by adding white noise and
a bias to its corresponding true value.

The local topocentric frame is defined as an ENU coordinate frame,
where the first, second, and third components correspond to East, North,
and Up, respectively. The rover position at epoch $k$ is denoted by
$\mathbf{x}_k=[x_k,y_k,z_k]^T$.

For position propagation, the navigation-sensor measurements at epoch
$k$ are corrected using the updated bias estimates from epoch $k-1$.
The bias estimates obtained in the measurement update at epoch $k$ are
used from the next prediction step onward.
The heading and vertical angle at epoch $k$ are predicted using the
measured angle increments corrected by the updated bias estimates from
the previous epoch as
\begin{equation}
\hat{\psi}_{k}^{-}
=
\hat{\psi}_{k-1}^{+}
+
\left(
\Delta\tilde{\psi}_{k}
-
\hat{b}_{\Delta\psi,k-1}^{+}
\right),
\end{equation}
and
\begin{equation}
\hat{\theta}_{k}^{-}
=
\hat{\theta}_{k-1}^{+}
+
\left(
\Delta\tilde{\theta}_{k}
-
\hat{b}_{\Delta\theta,k-1}^{+}
\right).
\end{equation}

The horizontal travel distance used for the prediction at epoch $k$
is corrected using the updated bias estimate from the previous epoch
as
\begin{equation}
\hat{d}_{H,k}
=
\tilde{d}_{H,k}
-
\hat{b}_{d_H,k-1}^{+}.
\end{equation}

The vertical displacement is estimated from the horizontal travel
distance and the predicted vertical angle as
\begin{equation}
\hat{d}_{V,k}
=
\hat{d}_{H,k}
\tan\hat{\theta}_{k}^{-}.
\end{equation}

Using the predicted heading, the three-dimensional relative
displacement is expressed as
\begin{equation}
\Delta\hat{\mathbf{x}}_k
=
\begin{bmatrix}
\hat{d}_{H,k}\cos\hat{\psi}_{k}^{-} \\
\hat{d}_{H,k}\sin\hat{\psi}_{k}^{-} \\
\hat{d}_{V,k}
\end{bmatrix}.
\end{equation}

The rover position is then propagated from the previous updated
position estimate as
\begin{equation}
\hat{\mathbf{x}}_{k}^{-}
=
\hat{\mathbf{x}}_{k-1}^{+}
+
\Delta\hat{\mathbf{x}}_k,
\label{eq:rover_position_propagation}
\end{equation}
where $\hat{\mathbf{x}}_{k}^{-}$ and
$\hat{\mathbf{x}}_{k}^{+}$ denote the predicted and updated rover
position estimates at epoch $k$, respectively.

Note that the directly measured heading $\tilde{\psi}_k$ and vertical
angle $\tilde{\theta}_k$ are not directly used in the position
propagation in Eq.~\eqref{eq:rover_position_propagation}. Instead, they
are used in the subsequent measurement update to correct the heading
and vertical-angle states and to estimate their corresponding
observation biases.

\subsubsection{Sensor Fusion Architecture}

The sensor-fusion KF combines the navigation-sensor-based state
propagation with an independent position solution obtained from the
radiometric positioning module. At each epoch, the navigation sensor
is first used to propagate the rover position, heading, and vertical
angle. When a valid radiometric position solution is available, it is
used as a position observation to update the sensor-fusion KF.

The navigation-sensor fusion KF estimates the rover position, heading,
vertical angle, and navigation-sensor biases. Its state vector is
defined as
\begin{equation}
\begin{aligned}
\mathbf{x}_{\mathrm{SF},k}
=
\big[
&\mathbf{x}_k^{T},\,
\psi_k,\,
\theta_k,\,
b_{d_H,k},\,
b_{\psi,k},\\
&b_{\theta,k},\,
b_{\Delta\psi,k},\,
b_{\Delta\theta,k}
\big]^{T}.
\end{aligned}
\end{equation}
where $\mathbf{x}_k$ is the rover position, $\psi_k$ is the horizontal
heading, $\theta_k$ is the vertical angle, $b_{d_H,k}$ is the
horizontal-distance bias, $b_{\psi,k}$ and $b_{\theta,k}$ are the
heading and vertical-angle observation biases, respectively, and
$b_{\Delta\psi,k}$ and $b_{\Delta\theta,k}$ are the corresponding
angle-increment biases.

To determine whether the measurement geometry is sufficiently favorable
for a radiometric position update, a VDOP-like vertical geometry metric
is calculated from the final-iteration Jacobian as
\begin{equation}
D_{V,k}
=
\sqrt{
\left[
\left(
\left(\mathbf{H}_{k}^{(5)}\right)^{T}
\mathbf{H}_{k}^{(5)}
\right)^{-1}
\right]_{3,3}
},
\label{eq:vdop_like_metric}
\end{equation}
where the third diagonal element corresponds to the rover Up component
in the ENU frame. Regardless of whether the DOWR-bias state is active,
$D_{V,k}$ is extracted from the element corresponding to the rover
vertical-position state. Because the Jacobian may include pseudorange,
weighted Doppler, and SS--rover DOWR observation rows with different
physical units and weighting, $D_{V,k}$ is referred to as a VDOP-like
metric rather than a conventional geometry-only VDOP.

A radiometric position update is accepted only when $D_{V,k}$ satisfies
the prescribed geometry threshold.


\section{Simulated Mission Scenario}

\subsection{Satellite Geometry}

\begin{figure*}[htbp]
    \centering
    \includegraphics[
        width=1\linewidth,
        trim={0 0 0 0},
        clip
    ]{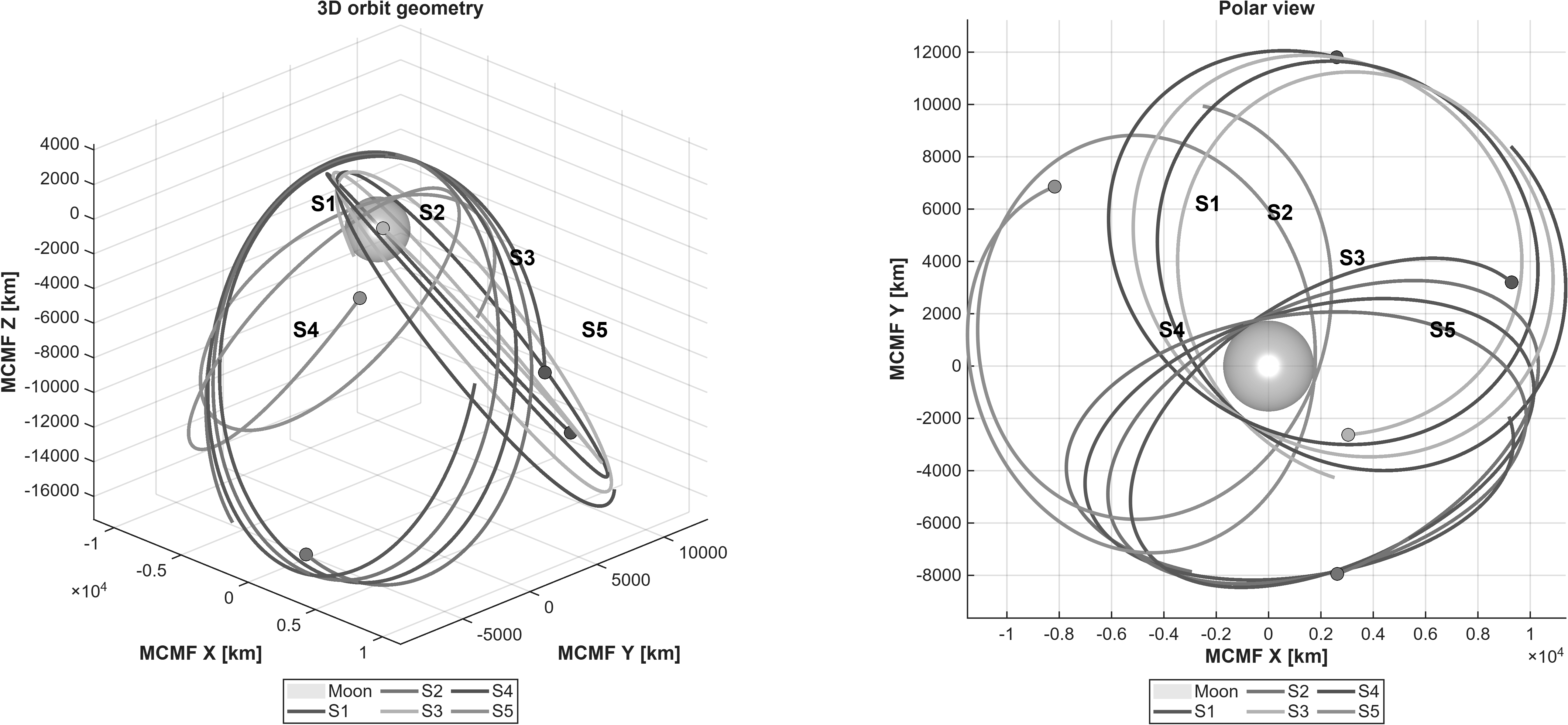}
    \caption{\footnotesize Simulated lunar navigation satellite orbits
    in the MCMF frame: three-dimensional view (left) and polar view
    (right).}
    \label{fig:satellite_orbits}
\end{figure*}

The lunar navigation satellite orbits used in this study were calculated
based on the constellation described in Ref.~\cite{ryden2025lcrns}.
The initial epoch of the satellite orbit propagation was set to
March 18, 2027, 06:00:00 Coordinated Universal Time (UTC).
Figure~\ref{fig:satellite_orbits} shows the simulated orbits of the
five satellites in the Moon-centered, Moon-fixed (MCMF) frame.

\subsection{SS Geometry}

\begin{figure}[htbp]
    \centering
    \includegraphics[
        width=1\linewidth,
        trim={{.0\linewidth} {.0\linewidth} {.0\linewidth} {.0\linewidth}},
        clip
    ]{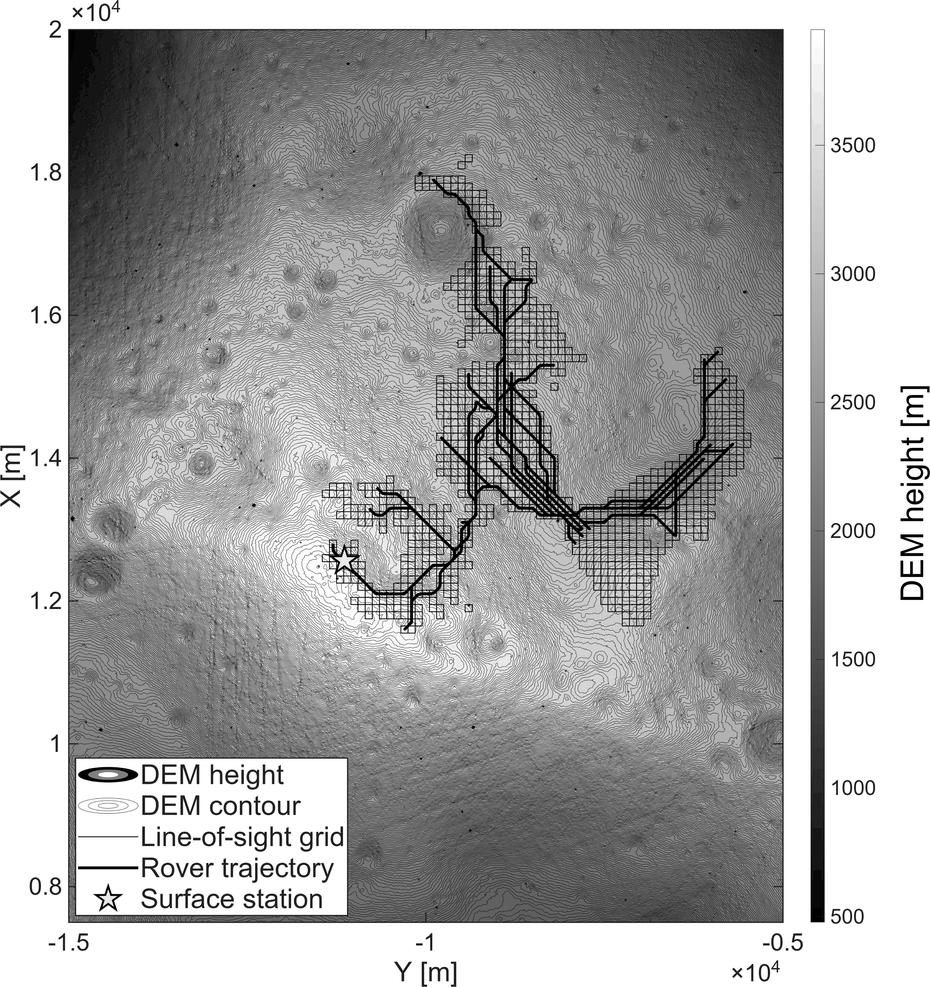}
    \caption{\footnotesize Simulated SS location, rover trajectory, and
    SS--rover LOS conditions. Each line-of-sight grid represents
    100~m $\times$ 100~m.}
    \label{fig:rover_trajectory}
\end{figure}

The SS was located at a latitude of $-89.45^\circ$ and an east
longitude of $221.56^\circ$. The Connecting Ridge region was selected
with reference to Ref.~\cite{gruener2020shackleton}, and the specific
SS location was defined in this study such that LOS with the rover
could be maintained over the surrounding exploration region, including
areas near permanently shadowed regions (PSRs), as shown in
Fig.~\ref{fig:rover_trajectory}.

\subsection{Rover Exploration Region and Trajectory}

The rover exploration region and trajectory were defined based on an
SS--rover LOS analysis.

\subsubsection{LOS Calculation}

The SS--rover LOS condition was evaluated using a DEM-based ray-tracing
procedure. The DEM product used in this study was the Lunar
Reconnaissance Orbiter/Lunar Orbiter Laser Altimeter (LRO/LOLA)
south-polar DEM \texttt{LDEM\_875S\_5M.JP2}, archived as part of the
LOLA GDR polar JP2 products. The DEM has a spatial resolution of 5~m
and covers latitudes from $-87.5^\circ$ to $-90^\circ$
\cite{LOLA_LDEM875S5M}.

The SS--rover LOS map was computed on a 100~m grid, obtained by
decimating the original 5~m DEM by a factor of 20. The SS and rover
antenna phase centers were assumed to be 8~m and 3~m above the DEM
surface, respectively. A clearance tolerance of 0.1~m was applied, and
each SS--rover ray was sampled at up to 100 intermediate points to
evaluate terrain blockage. The resulting LOS grid is shown as square
cells in Fig.~\ref{fig:rover_trajectory}.

The SS--rover DOWR observation was enabled only when LOS was available.
Note that the DEM was used to determine LOS availability but was not
used to provide an altitude constraint in the rover position estimation.

\subsubsection{LOS-Prioritized Waypoint Trajectory}

The rover trajectory was generated offline using 180 waypoints, all of
which were selected from LOS-labeled cells. The paths between waypoints
were generated to remain in LOS regions as much as possible, although
some portions of the connecting paths passed through NLOS cells. The
resulting rover path is shown in Fig.~\ref{fig:rover_trajectory}. This
resulted in LOS availability of 99.0\% for the grid-level path and
99.77\% for the final high-rate trajectory. The same precomputed
trajectory was used for all Monte Carlo runs.


\section{Numerical Simulations}

\subsection{Implementation Details}

The performance of the proposed navigation architecture was evaluated
through Monte Carlo simulations consisting of 30 runs for each case.

The simulation period was defined in terms of epochs, with one epoch
corresponding to 0.5~min. One set of radiometric observations was
generated at each epoch. The SS and rover observations were assumed to
be ideally synchronized.

The total simulation duration was set to 7200 epochs, corresponding to
3600~min, or approximately two orbital cycles of the lunar navigation
satellites. The first orbital cycle (epochs 1--3600) was used for DOWR
bias calibration and filter convergence. After the calibration period,
the final estimated DOWR bias was held fixed throughout the second
orbital cycle (epochs 3601--7200). Only the second orbital cycle was
used to evaluate the user positioning accuracy. This separation reduces
the influence of the initial filter transient and DOWR bias calibration
on the reported positioning performance.

For DOWR-bias estimation, the PDOP-like geometry metric $D_{P,k}$ was
used with a threshold of 3 to determine whether the observation geometry
was sufficiently favorable for the DOWR-bias update. For radiometric
position updates, the VDOP-like vertical geometry metric $D_{V,k}$ was
used with a threshold of 6. The Doppler weighting factor was set to 1. These threshold values were
selected empirically for the considered scenario; their generalization
to other trajectories and geometries remains to be investigated.

These parameters are summarized in Table~\ref{tab:simulation_parameters}.

{\setlength{\textfloatsep}{6pt}
\begin{table}[!t]
    \centering
    \caption{\footnotesize Simulation parameters.}
    \label{tab:simulation_parameters}
    \footnotesize
    \setlength{\tabcolsep}{3pt}
    \begin{tabularx}{\columnwidth}{@{}Xcc@{}}
        \hline
        Parameter & Value & Unit \\
        \hline
        Epoch interval & 0.5 & min \\
        Radiometric observation interval & 1 & epoch \\
        Total simulation period & 7200 & epochs \\
        Evaluation period & 3601--7200 & epochs \\
        \hline
        Doppler weighting factor & 1 & -- \\
        PDOP-like metric threshold for DOWR-bias update
            & 3 & -- \\
        VDOP-like metric threshold for radiometric position update
            & 6 & -- \\
        \hline
        Elevation cut & 15 & deg \\
        \hline
        Clock white noise, $\sigma_c$ & 1 & ns \\
        Clock bias, $b_c$ & $\pm 10$ & ns \\
        RAFS three-state clock coefficient, $\sigma_1$ & $1.0\times10^{-12}$ & -- \\
        RAFS three-state clock coefficient, $\sigma_2$ & $4.3\times10^{-17}$ & -- \\
        RAFS three-state clock coefficient, $\sigma_3$ & 0 & -- \\
        \hline
        Ephemeris error reset interval & 360 & epochs \\
        Ephemeris white noise (radial / along / cross)
            & 0.1 / 0.1 / 0.1 & m \\
        Ephemeris bias (radial / along / cross)
            & $\pm$2 / $\pm$10 / $\pm$10 & m \\
        Ephemeris random walk (radial / along / cross)
            & 0.5 / 2.5 / 2.5 & m/hr \\
        \hline
        Horizontal-distance noise / bias
            & 0.75 / 0.25 & m \\
        Heading-angle noise / bias
            & 0.2 / 0.1 & deg \\
        Vertical-angle noise / bias
            & 0.2 / 0.1 & deg \\
        Heading-increment noise / bias
            & 0.05 / 0.01 & deg \\
        Vertical-angle-increment noise / bias
            & 0.05 / 0.01 & deg \\
        Heading observation interval
            & 1 & epoch \\
        \hline
        Carrier frequency, $f_c$ & 2.2 & GHz \\
        Noise spectral density, $N_0$ & -204 & dB/Hz \\
        \hline
        Pseudorange transmit power, $W_{p,\mathrm{PR}}$ & 20 & W \\
        Pseudorange loop bandwidth, $B_n$ & 0.1 & Hz \\
        Pseudorange chip duration, $T_c$ & 977.5 & ns \\
        Pseudorange coherent integration time, $T_{\mathrm{CO}}$
            & 10 & ms \\
        \hline
        DOWR transmit power, $W_{p,\mathrm{DOWR}}$ & 1 & W \\
        DOWR loop bandwidth, $B_{\mathrm{stl}}$ & 1 & Hz \\
        DOWR chip duration, $T_r$ & 482 & ns \\
        DOWR hardware bias per TX/RX component
            & $U(0,\,4.5)$ & m \\
        \hline
        Rover and SS antenna gain & 0 & dBi \\
        Lunar navigation satellite antenna gain (S1--S5)
            & 10 & dBi \\
        \hline
    \end{tabularx}
\end{table}
}

\subsubsection{Satellite Observation Geometry Screening}

Ideally, terrain occlusion between the navigation satellites and rover
should be evaluated individually for each satellite at every epoch based
on the local terrain geometry. However,
performing such terrain-based satellite-to-rover LOS calculations
throughout the simulation would substantially increase the computational
cost. Therefore, a fixed
elevation cut was used as an approximation in this study.

\begin{figure}[htbp]
    \centering
    \includegraphics[
        width=\linewidth,
        trim={{.05\linewidth} {.05\linewidth} 0 0},
        clip
    ]{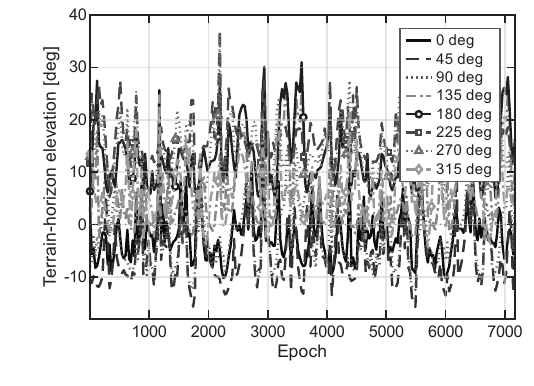}
    \caption{\footnotesize Terrain horizon elevation along the rover
    trajectory for eight azimuth directions.}
    \label{fig:terrain_horizon}
\end{figure}

To characterize the local terrain horizon, the maximum apparent terrain
elevation was evaluated along the rover trajectory in eight azimuth
directions, as shown in Fig.~\ref{fig:terrain_horizon}. An elevation cut
of $15^\circ$ was applied uniformly, which was estimated to screen more
than 95\% of local terrain occlusion caused by low satellite elevation.

\subsubsection{Clock Error Models}

The receiver and satellite clock errors, $\delta t_u$ and
$\delta t^{s_i}$, were generated independently using the clock model
adopted in Ref.~\cite{tanaka2026single}. For each clock $q$, the clock
offset relative to the reference time was modeled as
\begin{equation}
\delta t_q(t)=\omega_{c,q}(t)+b_{c,q}+d_{c,q}(t),
\label{eq:clock_error_model}
\end{equation}
where $\omega_{c,q}$, $b_{c,q}$, and $d_{c,q}$ denote independent white
clock noise, a bias component, and a time-varying clock drift, respectively.
The white noise was generated independently at each epoch, while the bias was
generated from a uniform distribution within the bounds listed in
Table~\ref{tab:simulation_parameters} and held constant for each run.
The drift component was generated using the three-state clock model of
Ref.~\cite{Zucca2005ClockModel}, using parameters representative of a Rubidium Atomic Frequency Standard
(RAFS). The clock parameters used in the simulation are
summarized in Table~\ref{tab:simulation_parameters}. Independent clock
errors were generated for the rover, SS, and each navigation satellite
and applied to the corresponding one-way pseudorange measurements.

\subsubsection{Satellite Ephemeris Error Implementation}

Satellite ephemeris errors were generated in the radial, along-track,
and cross-track directions following the error model in
Ref.~\cite{tanaka2026single}. The resulting errors were transformed
from the radial, along-track, and cross-track frame into the satellite
position-error vector $\delta\mathbf{x}_{\mathrm{eph}}^{s_i}$ for each
satellite $s_i$, which was then applied to the true satellite position
according to Eq.~\eqref{eq:satellite_ephemeris_error}.

In the present implementation, the ephemeris error in each radial,
along-track, and cross-track direction consisted of white Gaussian
noise, a piecewise-constant bias, and a random-walk component. The
white noise was generated independently at each epoch, while the bias
was generated from a uniform distribution within the bounds listed in
Table~\ref{tab:simulation_parameters} and held constant until the next
ephemeris-error reset. The random walk was propagated at each epoch
after converting the tabulated rate in m/hr to the corresponding
per-epoch value based on the simulation epoch interval.

Assuming a periodic orbit determination process, the satellite
ephemeris errors were reset every 360 epochs, corresponding to
180~min or 3~hr. At each reset, the piecewise-constant ephemeris
bias was re-sampled and the random-walk component was reset to zero.

\subsubsection{Navigation Sensor Error Implementation}

The navigation sensor was modeled by adding white Gaussian noise and a
constant random bias to each sensor measurement. For each sensor
channel, the bias was generated once per Monte Carlo run from a
zero-mean Gaussian distribution, using the bias value in
Table~\ref{tab:simulation_parameters} as the $1\sigma$ standard
deviation, and was held constant throughout that run. The white noise
was generated independently at each epoch using the standard deviation
listed in Table~\ref{tab:simulation_parameters}.

\subsubsection{Radiometric Error Models}

\paragraph{Pseudorange and Doppler Model}

For the satellite pseudorange observations, the receiver-noise
standard deviation was calculated as
\begin{equation}
\sigma_{r,\mathrm{PR}}
=
cT_c
\sqrt{
\frac{B_n}
{2(P_r/N_0)}
\left(
1+
\frac{2}
{T_{\mathrm{CO}}(P_r/N_0)}
\right)
},
\label{eq:pseudorange_receiver_noise}
\end{equation}
where $T_c$ is the pseudorange-code chip duration, $B_n$ is the
receiver noise bandwidth, and $T_{\mathrm{CO}}$ is the coherent
integration time.

The Doppler observations were generated by time-differencing the
noise-contaminated pseudorange observations. Specifically, the
pseudorange rate was obtained by differencing consecutive pseudorange
measurements in time, and the corresponding Doppler observation was
then calculated using Eq.~\eqref{eq:doppler_observation}. No additional
independent noise was added during this process. Temporal correlations
introduced by the time-differencing operation were neglected in the
present filter implementation.

For the satellite pseudorange receiver-noise model, the received
signal-to-noise-density ratio $P_r/N_0$ was calculated from the
free-space link budget as
\begin{equation}
\left(\frac{P_r}{N_0}\right)_{\mathrm{dBHz}}
=
10\log_{10}(W_p)
+
G_t
+
G_r
-
20\log_{10}
\left(
\frac{4\pi r f_c}{c}
\right)
-
N_0,
\label{eq:link_budget}
\end{equation}
where $W_p$ is the RF transmit power, $G_t$ and $G_r$ are the
transmit- and receive-antenna gains, respectively, $r$ is the
transmission distance, $f_c$ is the RF carrier frequency, and $N_0$
is the noise spectral density.

\begin{figure}[htbp]
    \centering
    \includegraphics[
        width=0.95\linewidth,
        trim={{.05\linewidth} {.05\linewidth} 0 0},
        clip
    ]{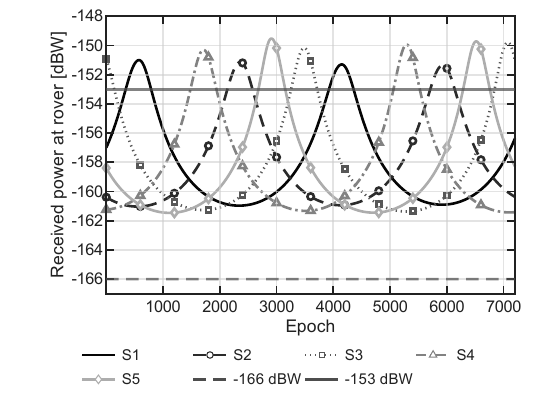}
    \caption{\footnotesize Received-power histories at the rover for
    lunar navigation satellites S1--S5.}
    \label{fig:received_power_history}
\end{figure}

The satellite transmit power and antenna gain used in this study were
determined based on the received-power levels specified in the
LunaNet Signal-In-Space Recommended Standard--Augmented Forward Signal
(LSIS) \cite{lunanet2023lsis}. The LSIS draft does not directly specify
the satellite transmit power or antenna gain assumed in this simulation.
Instead, the 2023 version of the LSIS specifies received-power levels
ranging from a minimum of $-166$~dBW to a maximum of $-153$~dBW.
Accordingly, the satellite transmit power and antenna gain were selected
such that the received power from each satellite satisfied the minimum
LSIS received-power level with an additional margin of approximately
3~dB. The resulting received-power histories for satellites S1--S5 are
shown in Fig.~\ref{fig:received_power_history}.

The resulting $P_r/N_0$ in dB-Hz was converted to its corresponding
linear value before being used in
Eq.~\eqref{eq:pseudorange_receiver_noise}. The receiver and
radiometric-link parameters used in the simulation are summarized in
Table~\ref{tab:simulation_parameters}.

\paragraph{DOWR Model}

For the SS--rover DOWR observation, the receiver-noise standard
deviation was calculated as
\begin{equation}
\sigma_{r,\mathrm{DOWR}}
=
cT_r
\sqrt{
\frac{B_{\mathrm{stl}}}
{2(P_r/N_0)}
\left(
1+\frac{3}{8R_s}
\right)
},
\label{eq:dowr_receiver_noise}
\end{equation}
where $T_r$ is the ranging-code chip duration,
$B_{\mathrm{stl}}$ is the loop bandwidth, and $R_s$ is the ranging
signal-to-noise ratio, defined as
\begin{equation}
R_s
=
\frac{P_r}{N_0 f_r},
\end{equation}
where
\begin{equation}
f_r
=
\frac{1}{T_r}
\end{equation}
is the range-clock waveform frequency.

The received signal-to-noise-density ratio $P_r/N_0$ for the DOWR
receiver-noise model was calculated from the free-space link budget
using Eq.~\eqref{eq:link_budget}. The resulting $P_r/N_0$ in dB-Hz
was converted to its corresponding linear value before being used in
Eq.~\eqref{eq:dowr_receiver_noise}.

The DOWR hardware bias $b_{\mathrm{DOWR}}$ was modeled as the sum of
four hardware-bias components corresponding to the transmitter and
receiver chains of the SS and rover. Each component was generated
independently as a constant random value from a uniform distribution,
and the resulting DOWR hardware bias was held constant throughout each
Monte Carlo run. The parameters used to generate the DOWR hardware bias
are summarized in Table~\ref{tab:simulation_parameters}.

\subsubsection{Error Statistics}
\label{sec:error_statistics}

As reference information, the statistics of the simulated error sources are presented below.

The statistics of the simulated satellite clock range error, satellite
ephemeris error along the satellite-to-rover LOS direction, and receiver
noise over the 30 Monte Carlo runs are summarized in
Table~\ref{tab:error_statistics}.
The values represent the mean and standard deviation of the 68\%, 95\%,
and 99.7\% error levels over the Monte Carlo runs.

While the satellite clock range errors and satellite ephemeris errors were
larger than the receiver noise, satellite clock errors can be eliminated
and common-mode components of satellite ephemeris errors can be mitigated
through DD processing using the SS observations. Therefore,
introducing the SS is expected to improve user positioning accuracy by
reducing the influence of these error sources.

\begin{table}[!htbp]
    \centering
    \caption{\footnotesize Statistics of the simulated error sources.}
    \label{tab:error_statistics}
    \scriptsize
    \setlength{\tabcolsep}{4pt}
    \renewcommand{\arraystretch}{1.08}
    \begin{tabular}{lccc}
        \hline
        \textbf{Satellite} & \textbf{68\%} & \textbf{95\%} & \textbf{99.7\%} \\
        \hline
        \multicolumn{4}{l}{\textit{Satellite clock range error [m]}} \\
        S1 & $2.0 \pm 0.39$ & $2.9 \pm 0.31$ & $3.4 \pm 0.32$ \\
        S2 & $2.0 \pm 0.33$ & $2.9 \pm 0.24$ & $3.4 \pm 0.26$ \\
        S3 & $2.0 \pm 0.34$ & $2.9 \pm 0.28$ & $3.4 \pm 0.28$ \\
        S4 & $2.2 \pm 0.34$ & $3.0 \pm 0.23$ & $3.5 \pm 0.23$ \\
        S5 & $2.0 \pm 0.35$ & $3.0 \pm 0.17$ & $3.5 \pm 0.19$ \\
        \hline
        \multicolumn{4}{l}{\textit{Satellite ephemeris error along satellite-to-rover LOS [m]}} \\
        S1 & $4.1 \pm 0.91$ & $7.6 \pm 2.0$ & $8.9 \pm 1.9$ \\
        S2 & $4.4 \pm 0.90$ & $8.2 \pm 1.8$ & $9.3 \pm 1.8$ \\
        S3 & $3.4 \pm 0.85$ & $5.8 \pm 1.1$ & $6.3 \pm 1.1$ \\
        S4 & $2.9 \pm 0.64$ & $5.8 \pm 1.4$ & $6.3 \pm 1.3$ \\
        S5 & $4.4 \pm 1.1$ & $7.7 \pm 1.5$ & $8.7 \pm 1.7$ \\
        \hline
        \multicolumn{4}{l}{\textit{Receiver noise [m]}} \\
        S1 & $0.33 \pm 0.01$ & $0.74 \pm 0.02$ & $1.2 \pm 0.05$ \\
        S2 & $0.33 \pm 0.01$ & $0.74 \pm 0.02$ & $1.2 \pm 0.05$ \\
        S3 & $0.35 \pm 0.01$ & $0.78 \pm 0.02$ & $1.3 \pm 0.06$ \\
        S4 & $0.35 \pm 0.01$ & $0.78 \pm 0.02$ & $1.3 \pm 0.05$ \\
        S5 & $0.35 \pm 0.01$ & $0.79 \pm 0.02$ & $1.3 \pm 0.05$ \\
        \hline
    \end{tabular}
\end{table}

\subsection{Simulation Results}

Two comparisons were performed. First, the contribution of the SS and
the additional benefit of SS--rover ranging were evaluated by comparing
three navigation cases: pseudorange and Doppler measurements without the
SS, DD PR/Doppler measurements using the SS without SS--rover ranging,
and DD PR/Doppler measurements with additional SS--rover ranging, i.e., DOWR. Second,
the effect of the residual DOWR hardware bias was evaluated by comparing
SS--rover ranging cases with DOWR-bias estimation disabled and enabled.

\subsubsection{Effect of the SS and SS--Rover Ranging}
\label{sec:ss_aided_ranging}

The effect of the SS was evaluated by comparing three navigation cases:
(1) pseudorange and Doppler measurements without use of the SS,
(2) DD pseudorange (DD PR) and Doppler measurements using
the SS but without SS--rover ranging, and (3) DD PR/Doppler measurements
with additional SS--rover DOWR. The first case represents a scenario in
which the SS is not used at all and therefore provides a baseline for
evaluating the overall benefit of introducing the SS. The comparison
between the second and third cases isolated the additional benefit of
SS--rover ranging. All other simulation conditions, including the
navigation-sensor error settings, were kept identical among the three
cases. For each Monte Carlo run, the same random seed was used for all
three cases so that identical navigation-sensor error realizations were
applied. The random seed was changed between Monte Carlo runs.

\begin{figure*}[htbp]
    \centering
    \includegraphics[
        width=0.9\linewidth
    ]{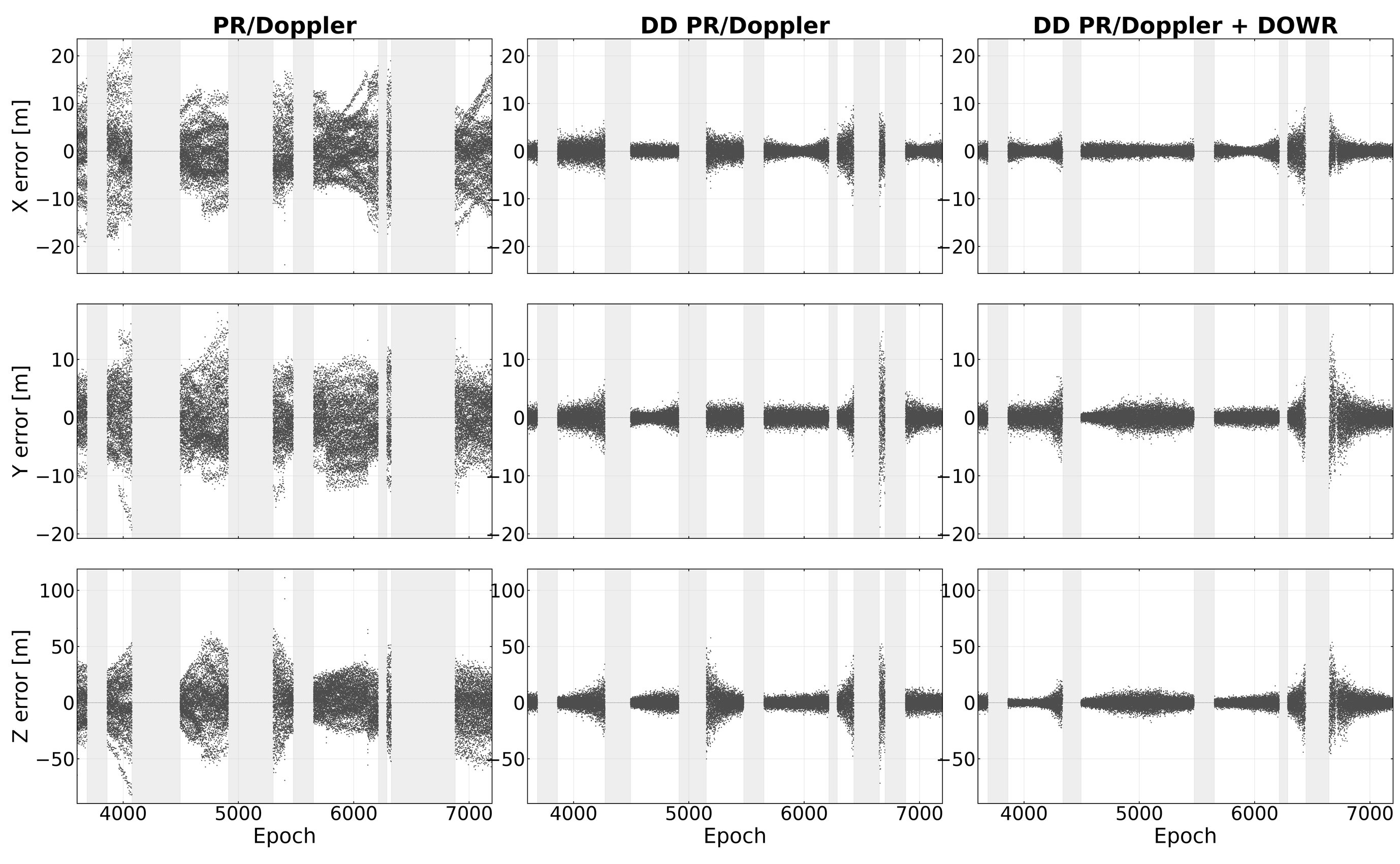}

    \vspace{1mm}

    \includegraphics[
        width=0.9\linewidth
    ]{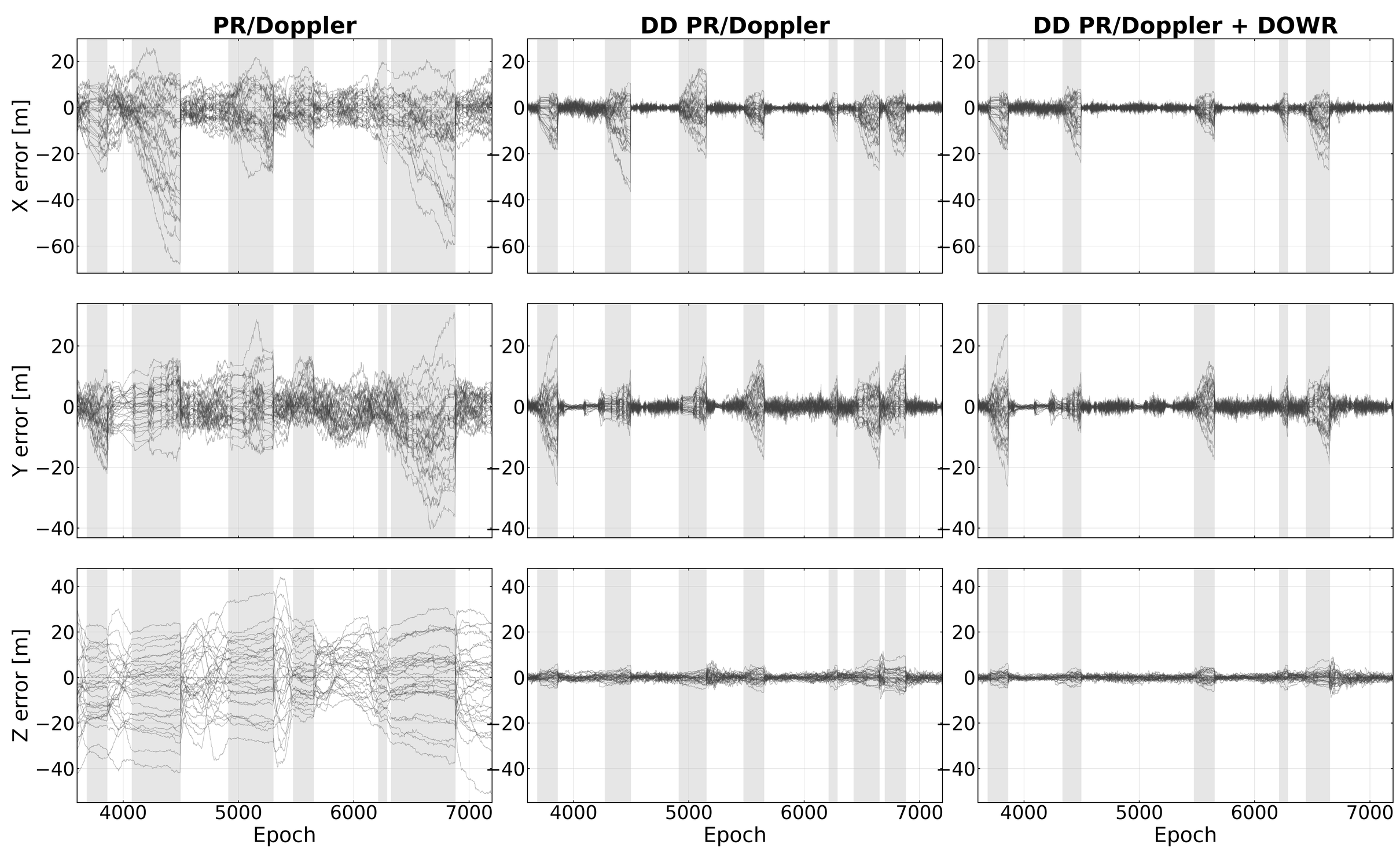}
    \caption{\footnotesize User positioning errors for the three navigation
    cases: pseudorange (PR) and Doppler without SS, DD PR/Doppler, and DD PR/Doppler with SS--rover DOWR.
    The upper panels show the radiometric position error
    $\delta\mathbf{x}_R$ obtained from the radiometric positioning without sensor fusion. The lower
    panels show the corresponding user positioning errors after integration
    with the navigation-sensor measurements through the sensor-fusion
    filter. The shaded regions indicate epochs in which radiometric
    position updates were unavailable. Each line represents one Monte
    Carlo run.}
    \label{fig:ss_three_case_positioning}
\end{figure*}

Figure~\ref{fig:ss_three_case_positioning} compares the rover positioning errors for the
three cases. The upper panels show the radiometric position error
$\delta\mathbf{x}_R$ obtained from the radiometric positioning without sensor fusion,
whereas the lower panels show the corresponding user positioning errors after
integration with the navigation-sensor measurements through the sensor-fusion filter. The shaded regions indicated epochs in which radiometric position updates were unavailable because the observation matrix was rank deficient or the PDOP-like metric $D_{P,k}$ exceeded the specified threshold. During periods without radiometric updates, the rover state was
propagated using only the navigation-sensor measurements; therefore,
navigation-sensor white noise and residual biases remaining after bias
estimation accumulated in the propagated rover state. 

\begin{figure*}[htbp]
    \centering
    \includegraphics[
        width=0.85\linewidth
    ]{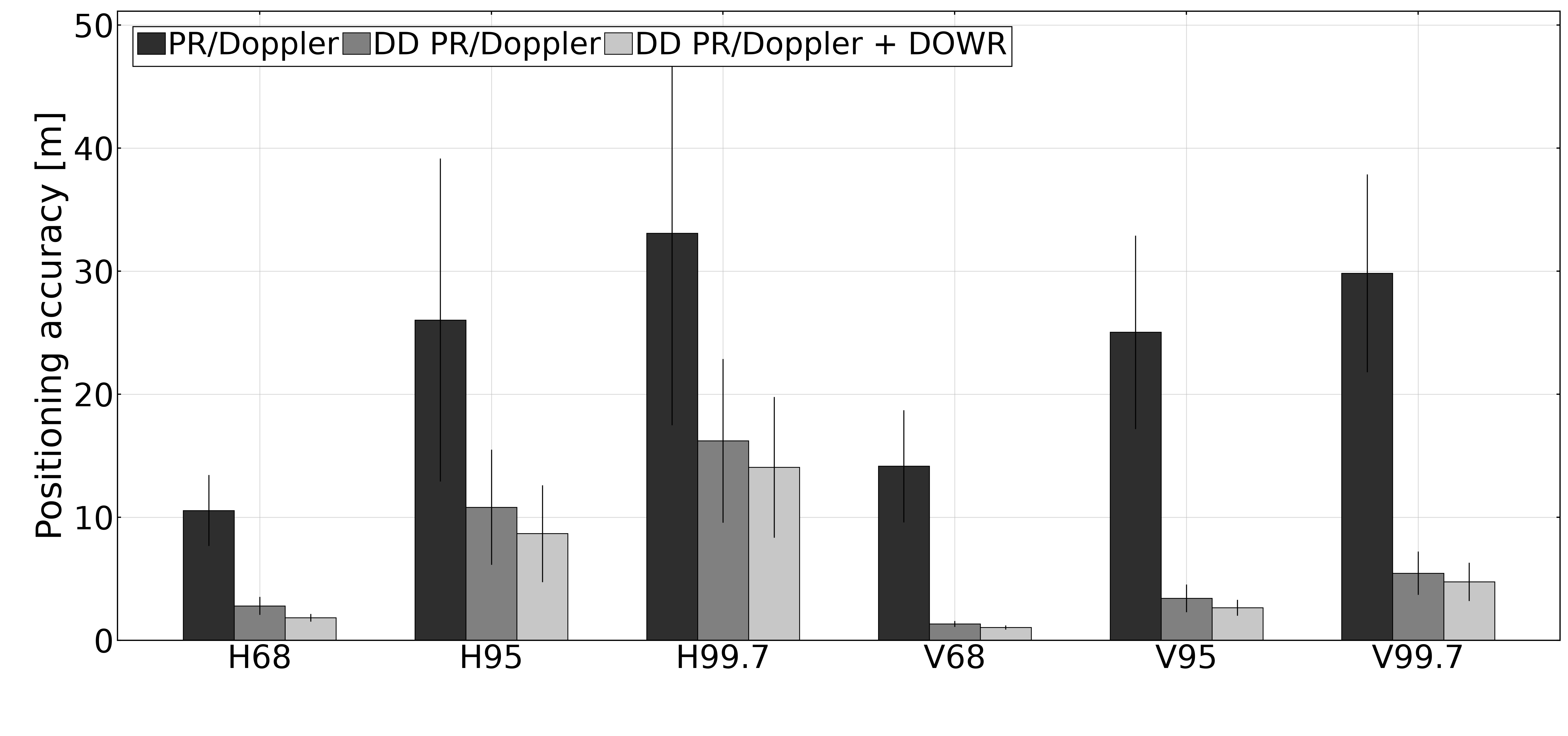}
    \caption{\footnotesize Visualization of the horizontal and vertical positioning accuracy for the three navigation cases summarized in Table~\ref{tab:ss_aided_positioning_accuracy}. Bars show the mean accuracy over the Monte Carlo runs, and error bars indicate the corresponding standard deviation.}
    \label{fig:ss_three_case_accuracy}
\end{figure*}

The horizontal and vertical positioning accuracies after integration with the
navigation-sensor measurements were then evaluated at the 68\%, 95\%, and
99.7\% percentile levels for each Monte Carlo run. The mean
and standard deviation of these values are summarized in
Table~\ref{tab:ss_aided_positioning_accuracy}, with the same results visualized
in Fig.~\ref{fig:ss_three_case_accuracy}. Introducing the SS
for DD PR/Doppler processing reduced both horizontal and
vertical errors, while the addition of SS--rover ranging provided a further
improvement in all evaluated metrics.


As shown in Table~3, the longest interval without a radiometric position update was shortest when SS--rover ranging was included. This reduction was attributed to the improved observation geometry provided by the additional SS--rover ranging measurement, which improved the PDOP-like metric \(D_{P,k}\) and consequently increased the number of epochs satisfying the geometry criterion for radiometric position updates.

\begin{table}[htbp]
    \centering
    \caption{\footnotesize Comparison of the user positioning performance after integration with the
    navigation-sensor measurements. Positioning errors are shown as mean $\pm$ standard
    deviation.}
    \label{tab:ss_aided_positioning_accuracy}
    \scriptsize
    \setlength{\tabcolsep}{1.5pt}
    \renewcommand{\arraystretch}{1.10}
    \begin{tabularx}{\columnwidth}{@{}
        >{\raggedright\arraybackslash}p{2.35cm}
        >{\centering\arraybackslash}X
        >{\centering\arraybackslash}X
        >{\centering\arraybackslash}X@{}}
        \hline
        \textbf{Metric}
        & \textbf{\scriptsize\shortstack{PR/Doppler\\
        (Without SS)}}
        & \textbf{\scriptsize\shortstack{DD PR/Doppler\\
        (Without\\
        SS--rover ranging)}}
        & \textbf{\scriptsize\shortstack{DD PR/Doppler\\
        + DOWR\\
        (With SS--rover\\
        ranging)}} \\
        \hline
        Horizontal accuracy, 68\% [m]
        & $10.6 \pm 2.9$ & $2.8 \pm 0.7$ & $1.8 \pm 0.3$ \\
        Horizontal accuracy, 95\% [m]
        & $26.1 \pm 13.1$ & $10.8 \pm 4.7$ & $8.7 \pm 3.9$ \\
        Horizontal accuracy, 99.7\% [m]
        & $33.1 \pm 15.6$ & $16.2 \pm 6.7$ & $14.1 \pm 5.7$ \\
        Vertical accuracy, 68\% [m]
        & $14.2 \pm 4.6$ & $1.3 \pm 0.2$ & $1.1 \pm 0.2$ \\
        Vertical accuracy, 95\% [m]
        & $25.1 \pm 7.9$ & $3.4 \pm 1.1$ & $2.7 \pm 0.7$ \\
        Vertical accuracy, 99.7\% [m]
        & $29.9 \pm 8.0$ & $5.5 \pm 1.8$ & $4.8 \pm 1.6$ \\
        Longest interval without radiometric update [epoch]
        & $555$ & $239$ & $207$ \\
        \hline
    \end{tabularx}
\end{table}

\paragraph{DOWR Hardware-Bias Estimation}

The convergence behavior of the DOWR hardware-bias estimate was
evaluated for the SS--rover ranging case.

\begin{figure*}[htbp]
    \centering
    \includegraphics[width=0.85\linewidth]{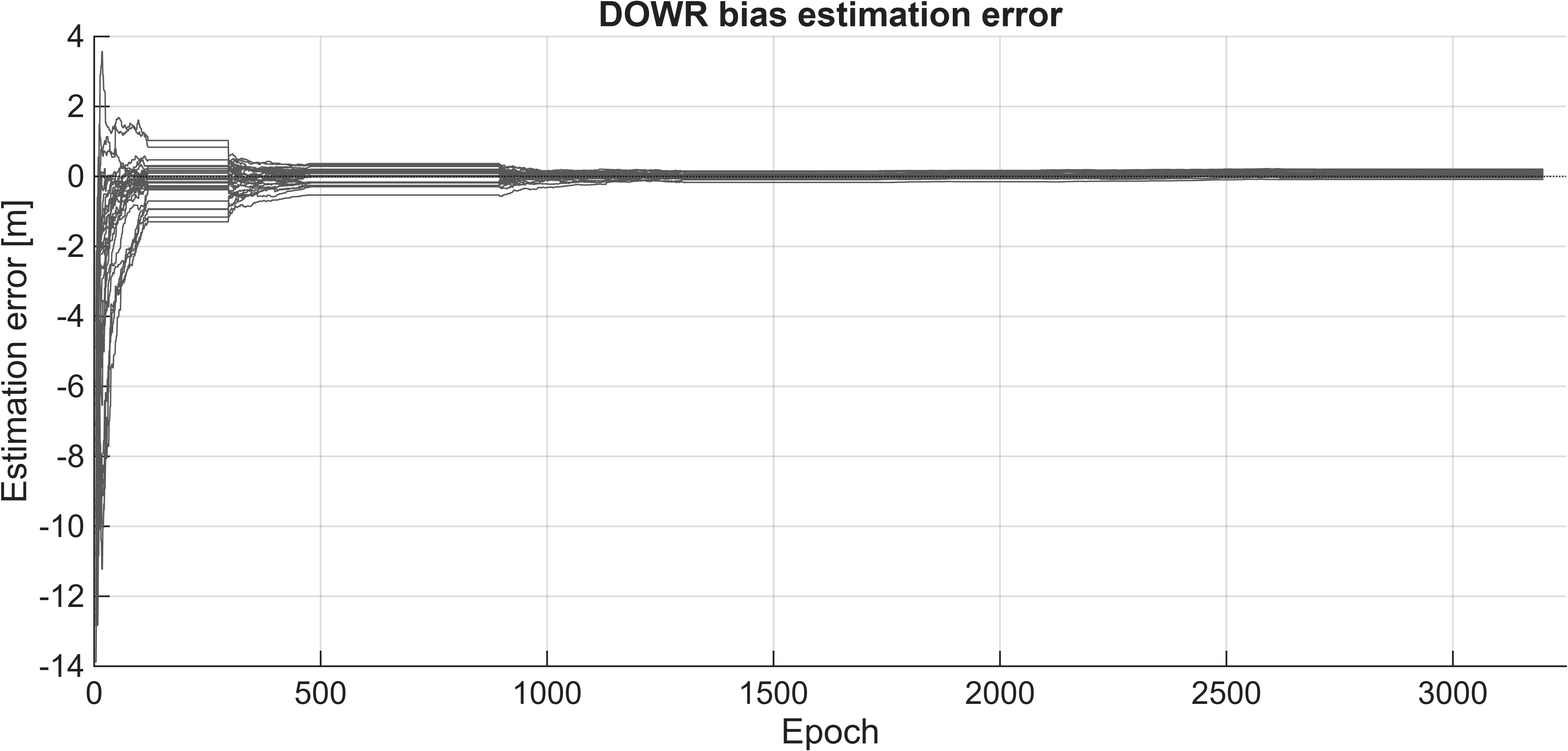}
    \caption{\footnotesize DOWR-bias estimation error during the
    calibration period. Each line represents one Monte Carlo run.
    The stepwise behavior of the DOWR-bias estimation error
    was caused by the geometry screening based on the PDOP-like metric
    $D_{P,k}$.}
    \label{fig:dowr_bias_convergence}
\end{figure*}

Figure~\ref{fig:dowr_bias_convergence} shows the DOWR-bias estimation
error during the calibration period. The DOWR-bias estimate converged
toward the true value within the first orbital cycle. The final DOWR-bias estimation
error after the calibration period was $0.058 \pm 0.057$~m.

The small final
mean error relative to the meter-level rover positioning errors
indicated that the residual DOWR bias after calibration was substantially
reduced by the proposed estimation process. These results indicate that, for the measurement geometry and simulation
conditions considered in this study, the DOWR-bias estimate sufficiently
converged within one orbital cycle.

\subsubsection{Effect of DOWR Hardware Bias}
\label{sec:dowr_hardware_bias_effect}

The effect of DOWR hardware-bias estimation was evaluated with
DOWR-bias estimation disabled and enabled, both in the SS--rover ranging scenario. This comparison isolates the importance of DOWR-bias
estimation within the SS--rover ranging architecture.

Table~\ref{tab:dowr_bias_sensitivity} summarizes the positioning
accuracy with and without DOWR-bias estimation. An uncorrected DOWR
hardware bias substantially degraded both horizontal and vertical
positioning performance.

\begin{table}[htbp]
    \centering
    \caption{\footnotesize Effect of DOWR-bias estimation on user
    positioning accuracy for two SS--rover ranging cases. Values are shown as
    mean $\pm$ standard deviation.}
    \label{tab:dowr_bias_sensitivity}
    \footnotesize
    \setlength{\tabcolsep}{2.0pt}
    \begin{tabularx}{\columnwidth}{@{}>{\raggedright\arraybackslash}p{3.8cm}>{\centering\arraybackslash}X>{\centering\arraybackslash}X@{}}
        \hline
        Metric & Bias est. OFF & Bias est. ON \\
        \hline
        Horizontal accuracy, 68\% [m]
            & $4.6 \pm 1.0$ & $1.8 \pm 0.3$ \\
        Horizontal accuracy, 95\% [m]
            & $10.7 \pm 3.3$ & $8.7 \pm 3.9$ \\
        Horizontal accuracy, 99.7\% [m]
            & $18.9 \pm 6.8$ & $14.1 \pm 5.7$ \\
        Vertical accuracy, 68\% [m]
            & $8.6 \pm 2.3$ & $1.1 \pm 0.2$ \\
        Vertical accuracy, 95\% [m]
            & $16.7 \pm 4.8$ & $2.7 \pm 0.7$ \\
        Vertical accuracy, 99.7\% [m]
            & $19.7 \pm 5.7$ & $4.8 \pm 1.6$ \\
        \hline
    \end{tabularx}
\end{table}


\section{Discussion}

\subsection{Key Takeaways}

The improvement associated with introducing the SS was particularly pronounced
in the vertical component. In Section 5.2.1, the mean 99.7\% vertical positioning error
decreased from 29.9~m with pseudorange and Doppler measurements alone to
5.5~m with DD PR/Doppler, and further to 4.8~m when SS--rover DOWR was
additionally used. The corresponding horizontal error decreased from
33.1~m to 16.2~m and then to 14.1~m.

Although DD processing and the SS--rover ranging improved both
components, its effect on the fused vertical accuracy was greater because
of the navigation architecture used in the simulation. Horizontal position
errors caused by heading errors can partially cancel as the rover changes
its direction of travel. In contrast, a constant vertical-angle bias tends
to produce vertical displacement errors with the same sign, causing them
to accumulate between radiometric updates. Therefore, the increased update
availability and shorter observation gaps provided by DD processing and SS--rover ranging more
effectively suppress navigation-sensor-induced drift in the vertical
direction.

At the same time, the
effectiveness of SS--rover ranging depends on appropriate treatment of
the DOWR hardware bias. As demonstrated in
Section~\ref{sec:dowr_hardware_bias_effect}, an uncorrected DOWR hardware
bias can significantly degrade positioning performance. Therefore,
accurate estimation and removal of this bias are important for achieving
the expected performance of the proposed architecture.

Note that the rover trajectory considered in this study was intentionally designed to prioritize SS--rover LOS availability. Therefore, the present results represent a favorable SS-visibility scenario, and performance under intermittent or sparse SS visibility remains to be investigated.

\subsection{Limitations}

Although SS--rover ranging improved the overall positioning performance
in the simulated scenario, from 5.5~m to 4.8~m as discussed in
Section~\ref{sec:ss_aided_ranging}, the magnitude of the
improvement varied over the course of the simulation periods.

\begin{figure*}[htbp]
    \centering
    \includegraphics[width=0.95\linewidth]
    {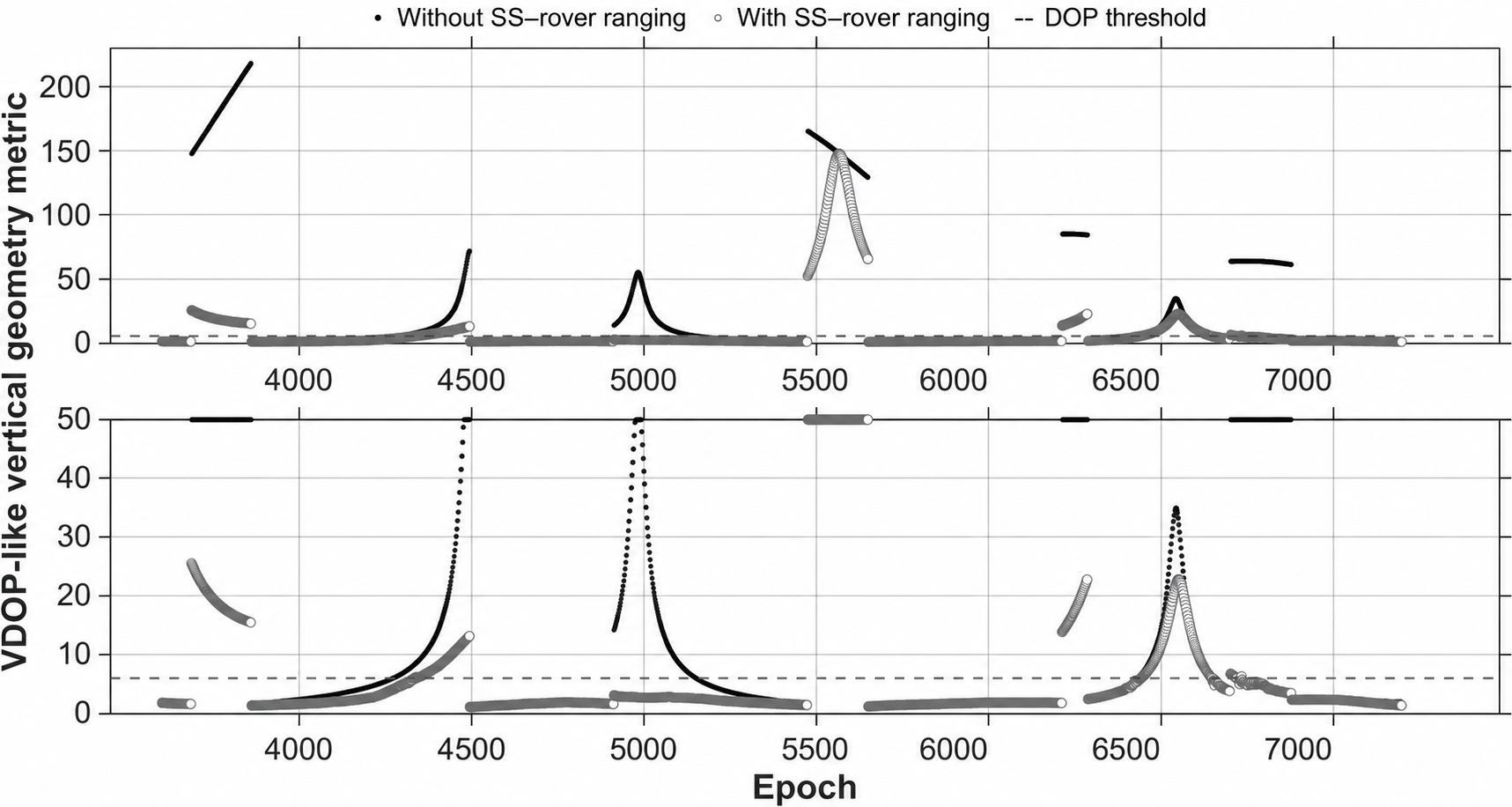}
    \caption{\footnotesize Comparison of the VDOP-like vertical geometry
    metric with and without SS--rover ranging. The upper and lower panels
    show the same data with different vertical-axis scales. The dashed line
    indicates the geometry threshold used for radiometric position updates.}
    \label{fig:dop_history}
\end{figure*}

\begin{figure*}[htbp]
    \centering
    \includegraphics[width=0.95\linewidth]
    {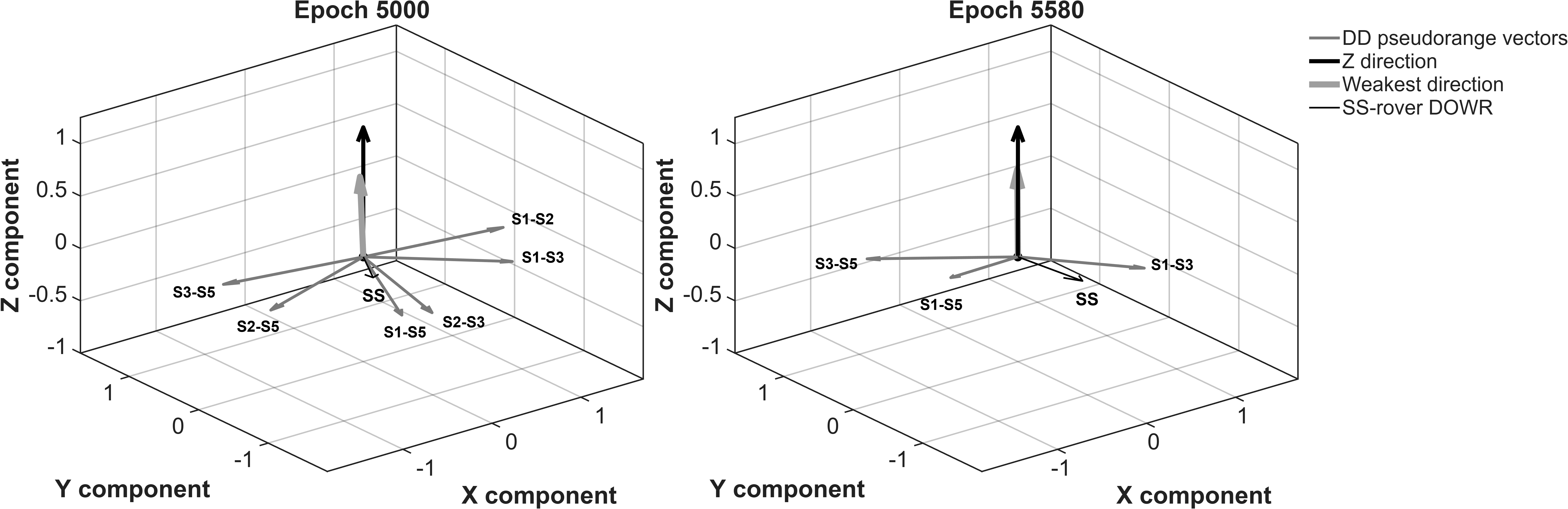}
    \caption{\footnotesize DD pseudorange observation geometry and
    SS--rover DOWR observation vector at epochs 5000 and 5580.}
    \label{fig:limited_improvements}
\end{figure*}

Figure~\ref{fig:dop_history} shows the VDOP-like vertical geometry
metric with and without SS--rover ranging. SS--rover ranging generally
improved the vertical observation geometry, but the magnitude of the
improvement varied significantly over time. Two representative epochs,
5000 and 5580, were examined in Fig.~\ref{fig:limited_improvements} to
illustrate this behavior. For simplicity, the effect of the DD Doppler
observations is omitted from the following geometric interpretation.
A more complete interpretation would also need to account for the
contribution of the DD Doppler observations.

At epoch 5000, four satellites (S1, S2, S3, and S5) were visible, and
six DD pseudorange pair directions are shown in
Fig.~\ref{fig:limited_improvements}. The VDOP-like vertical geometry
metric decreased from 41.16 without SS--rover ranging to 2.74 with
SS--rover ranging, corresponding to a 93.3\% improvement. The angle
between the SS--rover DOWR observation vector and the weakest direction
of the DD pseudorange observation geometry was $71.8^\circ$. Thus, the
SS--rover DOWR observation vector did not have a strong direct
projection onto the weakest direction. Nevertheless, the additional
SS--rover DOWR observation provided an independent geometric constraint
that was not available from the DD pseudorange observations alone.
This additional constraint improved the conditioning of the
overall observation geometry, resulting in a large reduction
in the VDOP-like vertical geometry metric.

In contrast, at epoch 5580, only three satellites (S1, S3, and S5)
were visible, and three DD pseudorange pair directions are shown in
Fig.~\ref{fig:limited_improvements}. The angle between the SS--rover
DOWR observation vector and the weakest direction was $89.6^\circ$,
indicating that the SS--rover DOWR observation provided almost no
direct constraint along the weakest direction. Consequently, adding
the SS--rover DOWR observation reduced the VDOP-like vertical geometry
metric only from 145.13 to 138.73, corresponding to an improvement of
4.4\%.

These results indicate that the positioning performance and the benefit
obtained from SS--rover ranging depend on the specific mission geometry and
navigation conditions. 
Consequently, the numerical results reported in this study are specific
to the simulated mission scenario, and the performance of the proposed
architecture should be evaluated under the specific conditions of a
target mission.

The present results also indicate that further analysis to understand
when SS--rover ranging is most effective would be useful. In particular,
the SS location and its geometry relative to the rover and navigation
satellites should be investigated to identify favorable configurations
for rover positioning.

\section{Conclusions}

This study investigated the use of surface stations (SS)-originated radiometric ranging specifically for improving the vertical positioning performance of lunar surface users. Previous studies have mainly used SS for differential corrections and satellite orbit determination, while some studies have also considered SS as additional ranging sources to improve the horizontal positioning accuracy of lunar surface users. Building on these previous approaches, this study demonstrated the additional value of SS-originated ranging for improving user \textit{vertical} positioning performance.

The proposed architecture consists of lunar navigation satellites, a SS, and a surface user. The SS receives pseudorange and Doppler signals from the navigation satellites, which are combined with the rover measurements to form double-differenced observations. In addition, dual one-way ranging (DOWR) is performed between the SS and rover. The rover combines the double-differenced satellite measurements, the SS--rover DOWR measurement, and its navigation-sensor measurements through filter-based estimation.

The main technical questions addressed in this study were whether an additional SS--rover ranging observation can improve vertical positioning performance, what the major error sources associated with the proposed architecture are and how they can be mitigated, and what factors limit the benefit provided by the SS. These aspects were quantitatively evaluated under a realistic lunar navigation scenario using a five-satellite LCRNS constellation and a SS located on Connecting Ridge near the lunar south pole.

The numerical results showed that introducing the SS substantially improved the rover positioning performance in the simulated scenario, with SS--rover ranging providing an additional benefit. In particular, the mean 99.7\% vertical positioning error was reduced from 29.9~m without the SS to 5.5~m when the SS was used for double-differenced pseudorange and Doppler processing, and further reduced to 4.8~m when SS--rover DOWR was additionally included. Although the additional improvement provided by SS--rover ranging was modest in the present scenario, the representative-case analysis showed that its benefit is strongly geometry dependent. More specifically, the benefit of SS--rover ranging can be limited when its line-of-sight direction provides little new geometric information relative to the existing radiometric measurements, suggesting that larger improvements may be achievable under more favorable SS--rover geometries.

One direction for further research is to generalize the conditions under which SS--rover ranging is especially effective through a systematic analysis of the observation geometry. In particular, the effect of SS geometry, including the SS location and the relative geometry between the SS, rover, and navigation satellites, will be evaluated to identify favorable configurations for improving rover positioning performance.

Overall, these results highlight an additional role that a SS can play beyond supporting communication, differential corrections, and satellite orbit determination. By providing an additional radiometric ranging signal directly to surface users, the same SS infrastructure can also contribute directly to improving rover positioning performance. This additional capability increases the potential value of SS as an integral part of an integrated lunar PNT and communication infrastructure.

\section*{Acknowledgements}

The authors gratefully acknowledge Yosuke Kawabata (Nagoya University)
for his technical support in generating the satellite orbit data.

\bibliographystyle{unsrt}
\bibliography{references}

\end{document}